\documentclass[runningheads]{llncs}

\usepackage[final,year=2026]{accv}
\usepackage{accvabbrv}
\usepackage{graphicx}
\usepackage{booktabs}
\usepackage{array}
\usepackage{multirow}
\usepackage{colortbl}
\usepackage{float}
\usepackage{placeins}
\usepackage{needspace}
\usepackage{amsmath}
\usepackage{amssymb}
\usepackage{algorithm}
\usepackage{algpseudocode}
\usepackage[accsupp]{axessibility}
\usepackage[breaklinks,colorlinks,citecolor=accvblue]{hyperref}
\usepackage{orcidlink}

\makeatletter
\renewcommand{\theALG@line}{\thealgorithm.\arabic{ALG@line}}
\providecommand{\theHALG@line}{\thealgorithm.\arabic{ALG@line}}
\renewcommand{\theHALG@line}{\thealgorithm.\arabic{ALG@line}}
\makeatother

\newcommand{\dataset}{MonoIR-RS}
\newcommand{\irclip}{IR-CLIP}
\newcommand{\irvlm}{IR-VLM}
\definecolor{oursrowgreen}{RGB}{232,245,226}
\definecolor{casegray}{RGB}{239,239,239}
\newcolumntype{L}[1]{>{\centering\arraybackslash}m{#1}}
\newcolumntype{C}[1]{>{\centering\arraybackslash}m{#1}}
\newcolumntype{G}[1]{>{\columncolor{casegray}\centering\arraybackslash}m{#1}}

\begin{document}
\raggedbottom

\title{\dataset: Infrared Remote Sensing Vision-Language Learning with CLIP and VLM Adaptation}
\titlerunning{\dataset}

\author{Jiaju Han\inst{1,3} \and Ma Yaqi\inst{2} \and Yahui Chai\inst{1} \and Xuemeng Sun\inst{1} \and Xin Li\inst{1} \and Qike Zhang\inst{1} \and Yingying Zhao\inst{1} \and Xiang Chen\inst{1} \and Luwei Yang\inst{3} \and Chengyin Hu\inst{1} \and Jiahuan Long\inst{4}}
\authorrunning{J. Han et al.}
\institute{China University of Petroleum-Beijing at Karamay, Karamay, Xinjiang, China
\and Guizhou University, Guiyang, China
\and Shenzhen Research Institute of Big Data, Shenzhen, China
\and Shanghai Jiao Tong University, Shanghai, China}

\maketitle

\begin{abstract}
Infrared remote-sensing imagery captures intensity structure, object-background contrast, and illumination-invariant cues often invisible in RGB imagery. Yet, most remote-sensing vision-language resources and models focus on visible-band semantics, leaving infrared vision-language understanding underexplored. We introduce \dataset{}, a large-scale infrared remote-sensing vision-language dataset and benchmark that couples IR-aware data construction with CLIP-style contrastive adaptation and VLM instruction tuning. Built from the same source pool and split as FusionRS, \dataset{} retains only the infrared image as the model-facing modality, yielding 600,000 synthesized infrared images and 59,032 retained IR-aware caption records. The model experiments use this retained language-supervision subset, whose captions rewrite supervision around grayscale structure and infrared-style contrast instead of RGB appearance. We show that the synthesized infrared is markedly closer to real thermal imagery than a grayscale conversion on the AVIID benchmark. We fine-tune five CLIP backbones and six VLM backbones, and calibrate them against zero-shot behavior: IR-aware adaptation lifts CLIP mean recall by up to $+12.8$ points (best checkpoint 19.2\% on the 9,720-image filtered split) and drives VLM captioning IR-cue coverage to 100\% while reducing residual RGB-color leakage to near zero. By isolating the infrared modality from RGB--IR dual-modal learning, \dataset{} offers a controlled, reproducible testbed for aligning infrared remote-sensing evidence with language.
\end{abstract}

\begin{figure}[t]
  \centering
  \includegraphics[width=\textwidth]{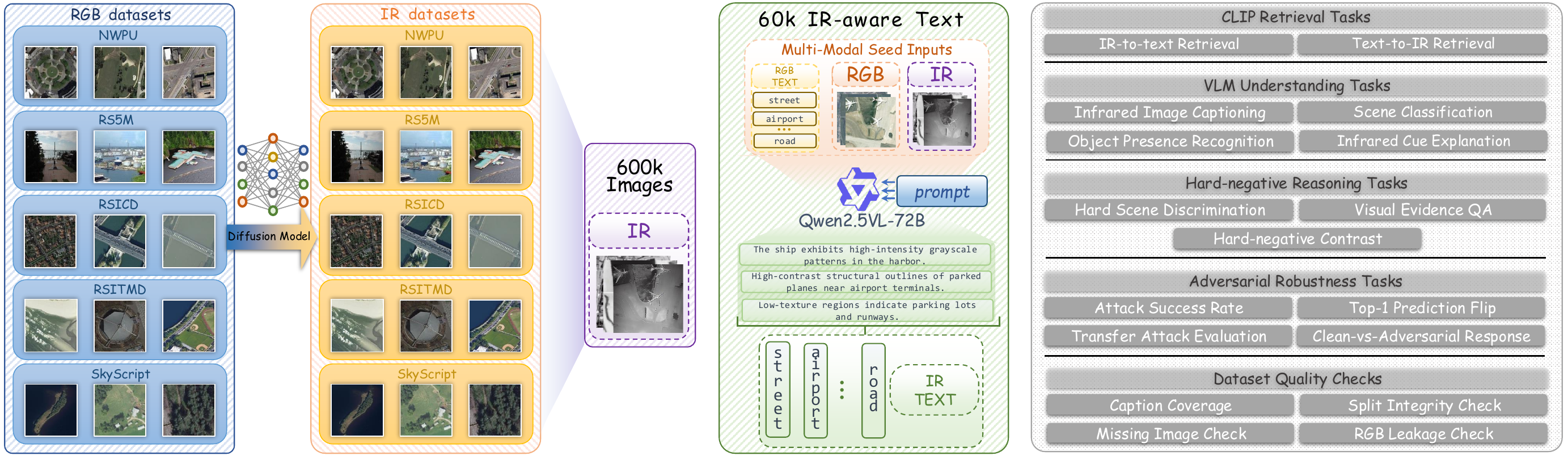}
  \caption{Overview of the \dataset{} construction workflow. The shared FusionRS source pool is converted into an infrared image-text corpus, IR-aware text, and evaluation tasks covering retrieval, VLM understanding, and dataset-quality checks.}
  \label{fig:teaser}
\end{figure}

\section{Introduction}
\label{sec:intro}

Infrared remote sensing is important for visual understanding under illumination changes, low-visibility conditions, and intensity contrast not captured by visible-band imagery, where discriminative evidence includes object-background separation and grayscale response rather than color. This makes infrared imagery valuable for nighttime monitoring, fire observation, and low-visibility analysis, yet difficult to serve with vision-language models trained on RGB web photography. The gap goes beyond a domain shift: infrared and visible sensors register the same scene under different physical principles, so captions written for visible images frequently describe color cues that are absent or misleading in the infrared modality.

Recent vision-language models, including CLIP-style contrastive models and instruction-following VLMs, have made remote-sensing retrieval, captioning, and question answering more practical. RemoteCLIP and GeoRSCLIP show that domain-specific supervision improves remote-sensing representations, while GRAFT aligns satellite and ground-level imagery without direct text annotations~\cite{liu2024remoteclip,zhang2024rs5m,mall2024remote}, and remote-sensing VLMs extend this toward grounded dialogue and geospatial benchmarking~\cite{kuckreja2024geochat,zhu2025skysense,danish2025geobench}. Infrared-specific efforts such as Infrared-LLaVA and IRGPT have begun to address the modality gap, but they target general surveillance or pedestrian infrared rather than remote sensing and rely on real sensor captures that are scarce and expensive to scale~\cite{jiang2024infrared,cao2025irgpt}. Transfer to infrared remote sensing therefore faces three bottlenecks: image-text resources are dominated by visible images; infrared captions must describe thermal evidence and grayscale structure rather than RGB appearance; and fair evaluation requires clean category protocols and strict split hygiene, since mixed phrases, residual RGB paths, or loose candidate pools can inflate conclusions.

This paper tackles infrared remote-sensing vision-language learning as an integrated data, adaptation, and evaluation problem. We build an IR-aware data pipeline that synthesizes infrared imagery from visible remote-sensing sources via diffusion and rewrites supervision around infrared-style visual evidence, fine-tune five CLIP and six VLM backbones under a train-only protocol, and evaluate against zero-shot behavior where the comparison is available. Rather than claiming photorealism, we validate the synthetic modality directly: it is closer to real thermal imagery than a grayscale conversion under both FID and histogram distance against the AVIID benchmark. As an auxiliary transfer sanity check, an IR-aware fine-tuned CLIP retrieves matching visible images from real infrared queries +6.5 mean-recall points better than its zero-shot counterpart on AVIID; this check is separate from the main infrared benchmark. This unified view (Fig.~\ref{fig:teaser}) lets us test whether IR-aware supervision benefits both retrieval models and VLMs under one data boundary, and whether gains survive controlled candidates, split audits, and per-source diagnosis. Our contributions follow.

\begin{itemize}
    \item We propose \dataset, a large-scale infrared remote-sensing vision-language dataset and benchmark with 600,000 synthetic infrared images and 59,032 retained IR-aware caption records supervised around infrared-style visual evidence. Validated against the real AVIID benchmark, its synthetic modality is closer to real infrared than a grayscale conversion by both FID and histogram distance.
    \item We adapt five CLIP and six VLM backbones under a train-only protocol, calibrating each family against zero-shot behavior. IR-aware CLIP adaptation yields $+3.5$ to $+12.8$ mean-recall gains (best 19.2\%), with a seed-stability check on OpenAI CLIP ViT-B/32, and transfers to real-infrared image retrieval on AVIID.
    \item We show that fine-tuning drives VLM captioning IR-cue coverage to 100\% while reducing RGB-color leakage relative to zero-shot models. Five diagnostic metrics---IR-cue rate, color-leakage rate, overclaim rate, class-hit rate, and response length---capture complementary failure modes that no single metric summarizes.
\end{itemize}

\section{Related Work}
\label{sec:related}

\subsection{Vision-Language Learning for Remote Sensing}
Remote-sensing vision-language learning has expanded from image classification toward retrieval, captioning, grounding, and multimodal reasoning, with CLIP-style contrastive learning aligning image-text embeddings and VLMs extending to free-form instructions. RemoteCLIP builds a remote-sensing foundation model from large-scale image-text supervision~\cite{liu2024remoteclip}; RS5M and GeoRSCLIP scale the dataset and model for retrieval and zero-shot recognition~\cite{zhang2024rs5m}; GRAFT aligns satellite with ground-level imagery to reduce annotation dependence~\cite{mall2024remote}; and longer-text and transductive methods further refine image-text matching~\cite{chen2025lrsclip,zanella2024boosting}.

\begin{sloppypar}
GeoChat studies grounded remote-sensing dialogue~\cite{kuckreja2024geochat}. SkySense and SkySense-O address open-world Earth observation interpretation~\cite{guo2024skysense,zhu2025skysense}. EarthDial, GEOBench-VLM, and large-image VLM work cover dialogue, benchmarking, and token handling~\cite{soni2025earthdial,danish2025geobench,luo2025large}. These efforts advance RGB and multi-sensor remote-sensing VLMs, but do not isolate the infrared setting, where color descriptions are unreliable and thermal evidence should guide supervision.
\end{sloppypar}

\subsection{Infrared and Thermal Remote-Sensing Data}
Infrared remote-sensing data differ from RGB in sensing physics and visual semantics: intensity contrast, sensor artifacts, and missing color make captions like ``green field'' or ``blue roof'' inappropriate for infrared scenes, so IR vision-language datasets need IR-aware text, not RGB captions copied onto infrared images. Table~\ref{tab:dataset_comparison} positions \dataset{} against nearby RGB--IR, infrared, and remote-sensing vision-language resources.

\begin{table}[tb]
  \centering
  \caption{Comparison with related datasets by modality, scale, domain, and primary use. The final row highlights the infrared vision-language setting of \dataset{}, in contrast to dual-modal RGB--IR resources and RGB-only remote-sensing image-text datasets.}
  \label{tab:dataset_comparison}
  \resizebox{\textwidth}{!}{%
  \begin{tabular}{@{}lllll@{}}
    \toprule
    Dataset & Modality & Scale & Domain & Primary use / limitation \\
    \midrule
    LLVIP~\cite{jia2021llvip} & RGB--IR & 15K pairs & Low-light surveillance & Detection / fusion; no language \\
    M3FD~\cite{liu2022m3fd} & RGB--IR & 4K pairs & Street scenes & Fusion / detection; non-RS \\
    MFNet~\cite{ha2017mfnet} & RGB--thermal & 1.6K pairs & Urban driving & Segmentation; no language \\
    VEDAI~\cite{razakarivony2016vedai} & Multi-spectral & 1.2K images & Aerial vehicles & Detection; no language \\
    DroneVehicle~\cite{sun2022dronevehicle} & UAV RGB--IR & 28K pairs & UAV traffic & Detection; no caption/VQA \\
    RSICD~\cite{lu2017rsicd} / RSITMD~\cite{yuan2021rsitmd} & RGB--text & 10.9K / 4.7K & Remote sensing & Caption/retrieval; RGB-only \\
    RS5M / GeoRSCLIP~\cite{zhang2024rs5m} & RGB--text & 5M pairs & Remote sensing & CLIP pretraining; RGB-only \\
    SkyScript~\cite{wang2024skyscript} & RGB--text & 2.6M pairs & Remote sensing + OSM & VLM data; RGB-only \\
    IRGPT / IR-TD~\cite{cao2025irgpt} & IR--text & 260K pairs & General infrared & IR language; non-RS \\
    FireMM-IR~\cite{cao2026firemmir} & RGB--IR + instr. & task-specific & Forest-fire RS & Dual-modal; scenario-specific \\
    FusionRS~\cite{han2026fusionrs} & RGB--IR--text & 600K triplets & General remote sensing & Dual-modal RGB--IR CLIP/VLM training \\
    \rowcolor{oursrowgreen}
    \dataset{} (ours) & IR--text + instr. & 600K IR images / 59K caption records & General remote sensing & Infrared CLIP/VLM fine-tuning \\
    \bottomrule
  \end{tabular}}
\end{table}
\FloatBarrier

Existing efforts span spectral and multi-modal Earth observation. S2MAE studies spatial-spectral pretraining for spectral RS~\cite{li2024s2mae}, AnySat targets multiple resolutions and modalities~\cite{astruc2025anysat}, and Infrared-LLaVA adapts MLLMs to infrared understanding~\cite{jiang2024infrared}. Most relevant, IRGPT~\cite{cao2025irgpt} builds a large real-infrared corpus and argues synthetic IR introduces a modality gap. We differ: IRGPT targets general IR (surveillance, pedestrian) with text produced by an LLM from the paired \emph{visible} image, whereas we target remote-sensing IR with supervision rewritten around infrared evidence; and we adopt DiffV2IR~\cite{ran2025diffv2ir}, whose visible-to-infrared translation attains strong fidelity (low FID/PSNR/SSIM against real thermal benchmarks), and validate the modality against real infrared directly (Sec.~\ref{sec:dataset}). Per the synthetic-data literature, we treat downstream utility, not photorealism, as the bar.

\subsection{Fine-Tuning CLIP and VLM Backbones}
Fine-tuning pretrained CLIP improves alignment for domain-specific image-text pairs, while VLM instruction tuning adapts multimodal assistants to specialized questions and scene descriptions. Domain prompt learning and controlled CLIP-transfer studies show that data distribution and adaptation strategy strongly affect downstream behavior, and that scaling data without quality control can degrade representations~\cite{cao2024domain,wen2024makes}. Visual instruction tuning is now the standard route for adapting VLMs~\cite{liu2024improved}, with work on data selection, compact visual tokens, and low-rank adaptation~\cite{safaei2025filter,zhang2025llava,bi2025llava}. These methods compare heterogeneous backbones under a shared training budget, which is essential when the contribution lies in data and protocol rather than one model.

For infrared remote sensing, CLIP and VLM fine-tuning play different roles: CLIP targets compact retrieval and embedding-space alignment, while VLMs target instruction following, captioning, and evidence-grounded QA. Our pipeline separates the two families and couples both with data-integrity diagnostics, so gains are judged by task scores and by whether they survive controlled candidate pools, split audits, and zero-shot calibration.

\begin{table}[t]
  \centering
  \caption{Distance to real infrared (AVIID, 804 test images). Synthetic infrared from DiffV2IR is compared with a direct RGB-to-gray baseline over Inception FID and grayscale-histogram metrics; lower is closer for FID/KL/JS/$\chi^2$, higher for histogram intersection.}
  \label{tab:ir_realism}
  \scriptsize
  \setlength{\tabcolsep}{5pt}
  \renewcommand{\arraystretch}{1.0}
  \begin{tabular}{@{}lccccc@{}}
    \toprule
    vs.\ Real IR & FID $\downarrow$ & KL $\downarrow$ & JS $\downarrow$ & $\chi^2$ $\downarrow$ & Hist.\ inter.\ $\uparrow$ \\
    \midrule
    Synthetic IR (DiffV2IR) & \textbf{85.2} & \textbf{0.103} & \textbf{0.027} & \textbf{0.051} & \textbf{0.822} \\
    RGB-to-gray baseline & 126.3 & 0.545 & 0.079 & 0.139 & 0.711 \\
    \bottomrule
  \end{tabular}
\end{table}

\section{Dataset Description}
\label{sec:dataset}

\subsection{Data Sources and Construction Goal}
\begin{figure}[t]
  \centering
  \includegraphics[width=\textwidth]{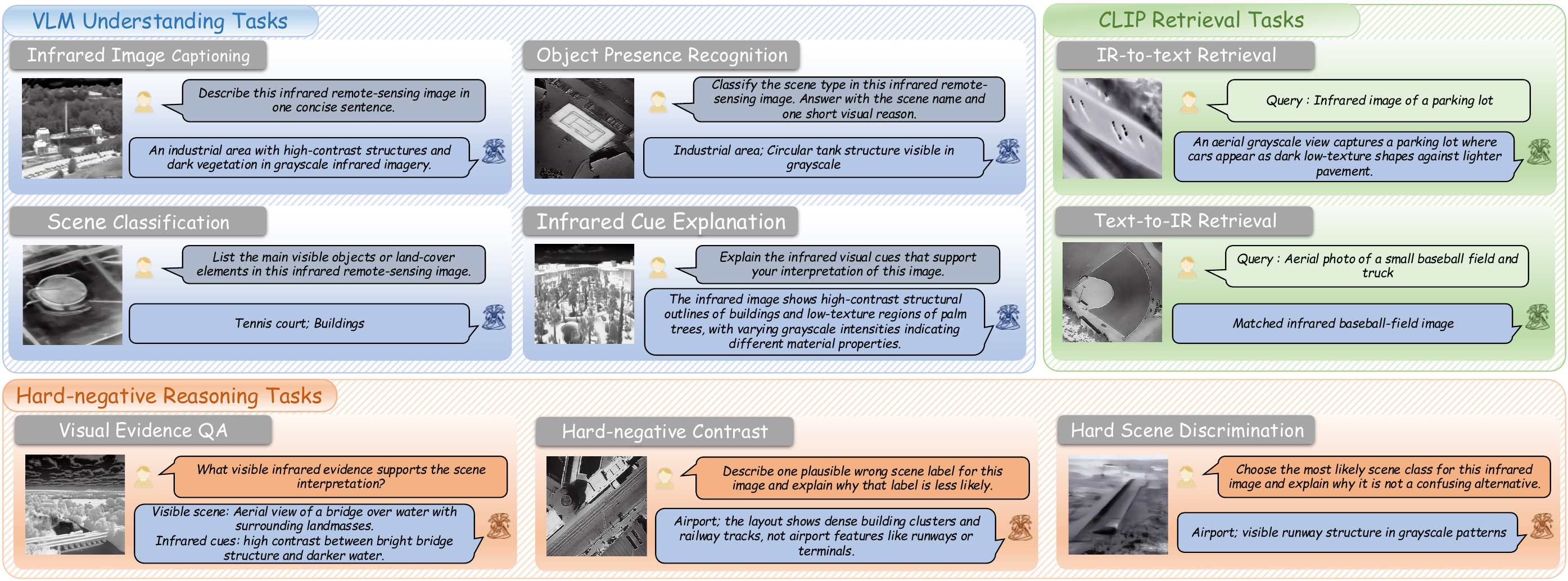}
  \caption{Representative \dataset{} task formats derived from infrared remote-sensing supervision, including VLM understanding, CLIP retrieval, and visual-evidence QA with paired prompts and target responses.}
  \label{fig:showcase}
\end{figure}

\dataset{} targets infrared remote-sensing vision-language training rather than dual-modal RGB--IR inference. It follows the same source pool and sample-level split as FusionRS but changes the modeling target: each sample exposes only the infrared image to the model, while the visible image is retained only for construction-time alignment and auditing. This isolates the infrared modality, where scene understanding must rely on grayscale structure, intensity contrast, and object-background separation rather than color. As shown in Tables~\ref{tab:dataset_statistics} and~\ref{tab:ir_caption_statistics}, the corpus contains 600,000 infrared samples from five sources and 59,032 IR-aware caption records after filtering. The image pool and the retained language-supervision subset are reported separately: not every synthesized infrared image contributes a retained caption after filtering, and the reported CLIP/VLM experiments use only retained train-split language supervision. RS5M and SkyScript provide large-scale coverage, while NWPU, RSICD, and RSITMD add established scene categories and fine-grained benchmark examples. The retained records are normalized into a unified infrared image-text format, supporting CLIP contrastive fine-tuning, VLM instruction tuning, and controlled evaluation under one split protocol. Figure~\ref{fig:showcase} previews the resulting task formats.

Because the five source collections provide only visible-band imagery, the infrared modality is synthesized rather than sensor-captured. We generate each infrared image from its registered visible image with DiffV2IR~\cite{ran2025diffv2ir}, a visible-to-infrared diffusion model that couples a progressive full-range-to-target-wavelength learning module with vision-language guidance and is trained on a large infrared image collection. This yields infrared-style renderings whose grayscale structure, object-background contrast, and thermal-response layout follow the source scene while removing color cues. We treat these renderings as a controlled, reproducible infrared testbed for vision-language adaptation rather than as radiometrically calibrated sensor measurements. This construction choice motivates our IR-aware supervision and RGB-leakage audits, since color words carried over from the source captions are unsupported by the synthesized infrared evidence.

To check that the synthesized modality is closer to real infrared than a naive grayscale conversion, we compare against the real aerial visible-infrared dataset AVIID~\cite{han2023aviid}. We render synthetic infrared from its 804 visible test images with the same DiffV2IR pipeline and compare the synthetic infrared, the real infrared, and a direct RGB-to-gray conversion (Table~\ref{tab:ir_realism}). By both FID and all four grayscale-histogram metrics, the synthetic infrared is markedly closer to real infrared than the grayscale baseline (FID 85.2 vs.\ 126.3). DiffV2IR thus captures infrared intensity structure beyond simple desaturation, though we still treat it as an infrared-like approximation, not a radiometrically calibrated sensor substitute.

Beyond distributional similarity, we run an auxiliary paired image-image retrieval sanity check on AVIID to test whether synthetic-IR adaptation transfers to real-infrared inputs. Using the selected IR-CLIP checkpoint, real-IR queries retrieve visible targets with 18.7\% mean recall, a $+6.5$-point gain over the corresponding zero-shot CLIP (12.2\%); with synthetic-IR queries against real-IR targets, the gain is $+23.4$ points. This experiment is not part of the main infrared benchmark itself, but it tests whether the learned infrared representation carries signal beyond the synthetic training domain.

\begin{table}[t]
  \centering
  \begin{minipage}[t]{0.49\textwidth}
    \centering
    \captionof{table}{Infrared split derived from the shared FusionRS source pool after retaining only model-facing IR images.}
    \label{tab:dataset_statistics}
    \scriptsize
    \setlength{\tabcolsep}{2.0pt}
    \renewcommand{\arraystretch}{1.0}
    \begin{tabular}{@{}lrrrr@{}}
      \toprule
      Source & Total & Train & Val & Test \\
      \midrule
      RS5M~\cite{zhang2024rs5m} & 488,033 & 471,729 & 8,166 & 8,138 \\
      SkyScript~\cite{wang2024skyscript} & 65,266 & 63,134 & 1,036 & 1,096 \\
      NWPU~\cite{cheng2017remote} & 31,186 & 30,122 & 549 & 515 \\
      RSICD~\cite{lu2017rsicd} & 10,824 & 10,494 & 166 & 164 \\
      RSITMD~\cite{yuan2021rsitmd} & 4,691 & 4,521 & 83 & 87 \\
      \midrule
      Total & 600,000 & 580,000 & 10,000 & 10,000 \\
      \bottomrule
    \end{tabular}
  \end{minipage}\hfill
  \begin{minipage}[t]{0.49\textwidth}
    \centering
    \captionof{table}{Distribution of IR-aware caption records across splits after filtering RGB-dependent, empty, and invalid entries.}
    \label{tab:ir_caption_statistics}
    \scriptsize
    \setlength{\tabcolsep}{3.0pt}
    \renewcommand{\arraystretch}{1.0}
    \begin{tabular}{@{}lrl@{}}
      \toprule
      Split & Records & Model use \\
      \midrule
      Train & 48,616 & CLIP training / VLM source pool \\
      Val & 416 & Checkpoint selection \\
      Test & 10,000 & Held-out evaluation \\
      \midrule
      Total & 59,032 & Train/val/test captions \\
      \bottomrule
    \end{tabular}
  \end{minipage}
\end{table}

\subsection{IR-Aware Text Generation}
Source captions often contain color words or visible-band details unreliable in infrared imagery, so we rewrite supervision to emphasize grayscale structure, thermal contrast, and scene layout. IR-aware captions are generated with Qwen2.5-VL-72B-Instruct~\cite{bai2025qwen25vltechnicalreport} from the source text and infrared evidence, preserving scene semantics while replacing RGB-dependent descriptions with infrared-grounded cues and discouraging unsupported color terms. Table~\ref{tab:ir_caption_statistics} summarizes the subset after filtering empty outputs, weak descriptions, unsupported claims, missing images, and invalid split entries. The retained supervision serves two formats: image-text pairs for CLIP contrastive fine-tuning and a train-only source pool from which VLM instruction-answer examples are constructed.

\subsection{Split Hygiene and Quality Control}
We constrain all fine-tuning data to the train split; validation and test images are excluded before CLIP or VLM training, and empty or non-train entries are treated as invalid. This shared boundary adapts both model families under the same train-only protocol, preventing hidden differences between the contrastive and instruction-tuning stages. We also keep dataset-integrity checks in the release surface: split-overlap auditing, missing-image checks, duplicate-path inspection, and RGB-leakage inspection, since residual visible-band paths or mixed entries can overstate infrared generalization. The original test split has 10,000 samples; the formal filtered surface retains 9,720 after flagging 280 RGB-named paths.

\section{CLIP and VLM Model Fine-Tuning}
\label{sec:finetune}

\subsection{Overview}
\begin{figure}[t]
  \centering
  \includegraphics[width=\textwidth]{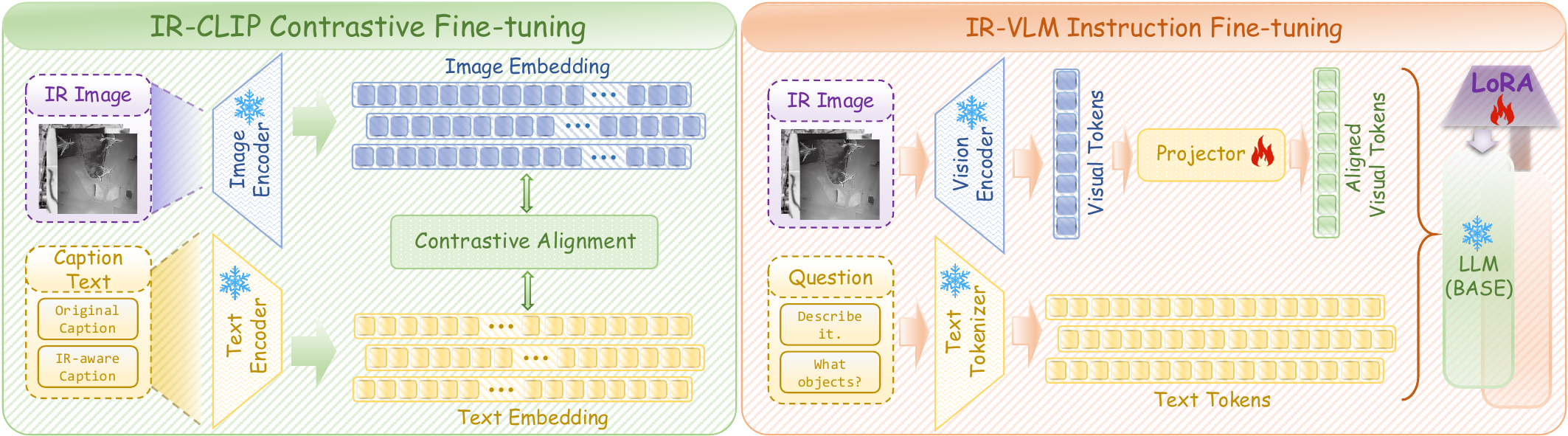}
  \caption{Two-stage model adaptation used for \dataset{}. The left branch fine-tunes CLIP-style encoders with infrared image-text contrastive alignment, while the right branch adapts VLM backbones with infrared visual tokens, a projector, and LoRA-based instruction tuning.}
  \label{fig:training}
\end{figure}

The model adaptation pipeline has two stages, as illustrated in Fig.~\ref{fig:training}. The first fine-tunes CLIP-style encoders for infrared-text contrastive alignment; the second adapts VLM backbones for instruction-style infrared understanding. We use fine-tuning rather than from-scratch pretraining to test how existing vision-language priors can be redirected toward infrared evidence under a controlled data boundary, keeping backbone capacity fixed so the effect of IR-aware supervision is comparable across models. The two stages are intentionally separated: CLIP provides compact embeddings for retrieval, whereas VLMs generate free-form answers for captioning, visual-evidence QA, object recognition, and infrared-cue explanation. This matters because retrieval alignment does not guarantee reliable instruction following, and fluent generation does not guarantee correct infrared grounding. Both stages use only train-split infrared images and IR-aware supervision.

\subsection{Training Data Interface}
The same infrared source boundary is exposed to the two model families through different interfaces. For CLIP, each retained train record is an image-text pair $(I_{\mathrm{ir}}, t_{\mathrm{ir}})$, matching the bidirectional retrieval objective used in evaluation. For VLMs, train-split records are converted and filtered into instruction triples $(I_{\mathrm{ir}}, q, a)$, where $q$ may request captioning, scene recognition, object presence, or infrared-cue explanation. The final VLM recipe uses the API-v3 multitask subset of 32,000 conversations selected from the train split, rather than all 48,616 retained train captions. RGB images stay outside the model-facing path, used only as construction-time alignment evidence. The two stages thus share one data discipline: train-only optimization and no use of held-out evaluation images during adaptation.

\subsection{IR-CLIP Contrastive Fine-Tuning}
For \irclip{} fine-tuning, the image encoder maps the infrared input into a visual embedding, the text encoder maps the IR-aware caption into a language embedding, and a symmetric contrastive objective aligns matched pairs while separating in-batch negatives. This adapts pretrained vision-language priors to infrared remote sensing without relying on RGB appearance, where the visual evidence is dominated by grayscale layout, thermal contrast, sensor-specific texture, and fine structural boundaries. We evaluate five CLIP backbones spanning general and remote-sensing variants. During fine-tuning, the text side is frozen and the visual tower plus projection layers are adapted, so the main optimization pressure falls on aligning infrared evidence to the existing language space while preserving stable text semantics. Checkpoints are selected by validation mean recall and used for bidirectional IR-text retrieval.

\subsection{IR-VLM Instruction Fine-Tuning}
For \irvlm{} fine-tuning, each example contains an infrared image, an instruction, and a target answer derived from IR-aware supervision. The vision encoder extracts infrared visual tokens, a projector maps them into the language-model embedding space when the backbone exposes a trainable projector, and LoRA adapters update instruction-following behavior while keeping the base language model frozen. This parameter-efficient setting makes it feasible to compare heterogeneous VLM backbones under a common infrared interface and reduces confounding from full-model optimization. The instruction data cover complementary forms of infrared understanding: captioning prompts summarize global scene structure, scene and object prompts encourage short semantic decisions, and infrared-cue or visual-evidence prompts require answers grounded in grayscale layout, object-background contrast, thermal evidence, and observable scene geometry.

\begin{table}[t]
  \centering
  \caption{Training configuration for \irclip{} and \irvlm{} fine-tuning. Backbones are listed together within each family, while shared trainable modules, objectives, schedules, and checkpoint-selection rules are reported once per family.}
  \label{tab:training_config}
  \scriptsize
  \setlength{\tabcolsep}{5pt}
  \renewcommand{\arraystretch}{1.25}
  \begin{tabular}{@{}>{\raggedright\arraybackslash}p{0.14\textwidth} >{\raggedright\arraybackslash}p{0.80\textwidth}@{}}
    \toprule
    \textbf{Field} & \textbf{Configuration} \\
    \midrule
    \multicolumn{2}{@{}l@{}}{\cellcolor{oursrowgreen}\textbf{\irclip{} family} --- contrastive IR image-text fine-tuning} \\
    \cmidrule(l){1-2}
    Backbones & OpenAI CLIP ViT-B/32, OpenAI CLIP ViT-L/14, OpenCLIP ViT-B/32, RemoteCLIP ViT-B/32, GeoRSCLIP ViT-B/32 \\
    \cmidrule(l){1-2}
    Trainable / objective & Visual tower and projection layers; text encoder frozen; contrastive IR image-text alignment \\
    \cmidrule(l){1-2}
    Schedule & 2 epochs; batch size 768 (OpenAI B/32), 128 (OpenAI L/14), 1024 (OpenCLIP, RemoteCLIP, GeoRSCLIP); learning rate $2{\times}10^{-6}$ for B/32-family backbones and $1{\times}10^{-6}$ for OpenAI L/14 \\
    \cmidrule(l){1-2}
    Selection & Best checkpoint by validation mean recall \\
    \midrule
    \multicolumn{2}{@{}l@{}}{\cellcolor{oursrowgreen}\textbf{\irvlm{} family} --- instruction-style infrared understanding} \\
    \cmidrule(l){1-2}
    Backbones & Qwen2.5-VL-7B~\cite{bai2025qwen25vltechnicalreport}, InstructBLIP-FLAN-T5-XL, LLaVA-1.5-7B~\cite{liu2024improved}, LLaVA-1.6-Vicuna-7B, GeoChat-7B~\cite{kuckreja2024geochat}, H2RSVLM-VHM-7B~\cite{pang2025vhm} \\
    \cmidrule(l){1-2}
    Trainable / objective & LoRA adapters ($r=32$, $\alpha=64$, dropout 0.05) and trainable multimodal projector where exposed by the backbone; base LLM frozen; instruction-following loss \\
    \cmidrule(l){1-2}
    Schedule & 1 epoch; batch/accum.\ 1$\times$8 (Qwen), 2$\times$8 (InstructBLIP), 4$\times$4 (LLaVA, GeoChat, H2RSVLM); learning rate $2{\times}10^{-4}$, cosine schedule, 0.03 warmup ratio \\
    \cmidrule(l){1-2}
    Selection & Final-epoch adapter; save-total-limit 1 \\
    \bottomrule
  \end{tabular}
\end{table}

\subsection{Training Configuration}
Table~\ref{tab:training_config} reports the training configuration for the two model families: CLIP uses contrastive alignment with a frozen text encoder, and VLMs use parameter-efficient instruction tuning with LoRA adapters and projector tuning where applicable. The settings keep each family within a standard regime while preserving the same data boundary and evaluation protocol. CLIP checkpoints are selected by validation mean recall; for VLMs we retain the final-epoch adapter under a fixed one-epoch schedule. All fine-tuning and evaluation are restricted to a single GPU by setting \texttt{CUDA\_VISIBLE\_DEVICES=0}, and each configuration is trained once; we therefore report single-run results, plus a three-seed stability check for OpenAI CLIP ViT-B/32 in Sec.~\ref{sec:clip_results}.

\section{Experiments}
\label{sec:experiments}

\subsection{Evaluation Protocol}
The formal CLIP evaluation uses the filtered infrared test split with 9,720 samples after auditing the original 10,000-image test split, and all reported evaluation samples are excluded from fine-tuning. VLM diagnostics use held-out infrared prompts and report task-level averages over the six final VLM backbones; the prompt sampling files are part of the release artifacts. We separate the evaluation by model family because each measures a different aspect of infrared vision-language learning: retrieval evaluates whether infrared images and IR-aware captions are aligned in a shared embedding space, whereas VLM diagnostics evaluate whether an instruction-tuned model produces language grounded in infrared evidence. Combining these into a single aggregate would obscure family-specific failure modes.

\subsection{IR-CLIP Retrieval}
\label{sec:clip_results}

\begin{table}[t]
  \centering
  \caption{Clean IR-CLIP retrieval on the 9,720-image filtered infrared test split. All values are percentages; mR averages R@1, R@5, and R@10 for both retrieval directions. The final two columns compare zero-shot mR (original pretrained weights, no IR-aware fine-tuning) with our fine-tuned mR, and $\Delta$ is the absolute gain from IR-aware adaptation.}
  \label{tab:clip_retrieval}
  \scriptsize
  \setlength{\tabcolsep}{3.0pt}
  \renewcommand{\arraystretch}{1.0}
  \resizebox{\textwidth}{!}{%
  \begin{tabular}{@{}lcccccc|ccc@{}}
    \toprule
    Model & I2T R@1 & I2T R@5 & I2T R@10 & T2I R@1 & T2I R@5 & T2I R@10 & Zero-shot mR & mR & $\Delta$ \\
    \midrule
    OpenAI CLIP ViT-L/14 & \textbf{8.3} & \textbf{19.8} & \textbf{27.1} & \textbf{9.5} & \textbf{21.4} & \textbf{28.8} & 6.3 & \textbf{19.2} & \textbf{+12.8} \\
    GeoRSCLIP ViT-B/32 & 5.3 & 14.1 & 20.4 & 5.6 & 14.2 & 20.6 & 4.2 & 13.4 & +9.2 \\
    OpenCLIP ViT-B/32 & 5.1 & 13.0 & 18.5 & 6.2 & 14.1 & 19.8 & 5.5 & 12.8 & +7.3 \\
    OpenAI CLIP ViT-B/32 & 4.8 & 12.5 & 17.7 & 5.5 & 13.0 & 18.5 & 4.3 & 12.0 & +7.7 \\
    RemoteCLIP ViT-B/32 & 1.1 & 4.0 & 6.5 & 1.5 & 4.7 & 7.8 & 0.8 & 4.3 & +3.5 \\
    \bottomrule
  \end{tabular}}
\end{table}

For \irclip{} fine-tuning, the image encoder maps the infrared input into a visual embedding, the text encoder maps the IR-aware caption into a language embedding, and a symmetric contrastive objective aligns matched pairs while separating in-batch negatives. This adapts pretrained vision-language priors to infrared remote sensing without relying on RGB appearance, where the visual evidence is dominated by grayscale layout, thermal contrast, sensor-specific texture, fine structural boundaries, and object-background separation. We evaluate five CLIP backbones spanning general and remote-sensing variants to test whether infrared adaptation benefits both generic and geospatially pretrained representations. During fine-tuning, the text side is frozen and the visual tower plus projection layers are adapted, so the main optimization pressure falls on aligning infrared evidence to the existing language space while preserving stable text semantics. This design avoids re-learning the caption space and instead asks the visual encoder to reinterpret infrared patterns as language-grounded remote-sensing concepts. Checkpoints are selected by validation mean recall and used for bidirectional IR-text retrieval.

\begin{table}[t]
  \centering
  \caption{Per-source clean retrieval on the filtered infrared test split, retrieving within each source's own candidate pool. mR averages R@1, R@5, and R@10 over both directions. Small pools (RSICD, RSITMD) inflate recall and are not directly comparable across sources; sample counts are reported for context.}
  \label{tab:clip_persource}
  \tiny
  \setlength{\tabcolsep}{2.8pt}
  \renewcommand{\arraystretch}{0.92}
  \scalebox{0.92}{%
  \begin{tabular}{@{}lrcc@{}}
    \toprule
    Source & \# Samples & OpenAI ViT-L/14 mR & GeoRSCLIP ViT-B/32 mR \\
    \midrule
    RS5M & 7,858 & 22.8 & 15.8 \\
    SkyScript & 1,096 & 10.8 & 10.8 \\
    NWPU & 515 & 34.0 & 31.7 \\
    RSICD & 164 & 37.7 & 32.6 \\
    RSITMD & 87 & 54.6 & 54.6 \\
    \bottomrule
  \end{tabular}
  }
\end{table}

Table~\ref{tab:clip_retrieval} reports clean bidirectional retrieval on the formal filtered test split. OpenAI CLIP ViT-L/14 achieves the strongest result, reaching 19.2\% mean recall and leading both retrieval directions across all recall cutoffs. GeoRSCLIP and OpenCLIP form the next tier, while OpenAI CLIP ViT-B/32 remains competitive despite lacking remote-sensing-specific pretraining. RemoteCLIP performs poorly under this infrared split despite remote-sensing adaptation, suggesting that RGB-oriented remote-sensing pretraining does not transfer uniformly when evidence is dominated by grayscale layout, thermal contrast, and sensor-specific texture.

The task remains challenging: even the best model reaches only 8.3\% I2T R@1 and 9.5\% T2I R@1, since many remote-sensing categories share similar infrared layouts and removing color eliminates shortcuts used by RGB-trained models. The consistent gain of ViT-L/14 over B/32 variants suggests stronger visual capacity helps capture infrared structure under IR-relevant supervision. To calibrate these numbers, the last three columns of Table~\ref{tab:clip_retrieval} compare each backbone against its own zero-shot performance using the original pretrained weights without IR-aware fine-tuning. Zero-shot mean recall is low across all backbones (0.8--6.3\%), confirming that off-the-shelf CLIP transfers poorly to synthetic infrared imagery, and IR-aware adaptation yields a consistent absolute gain of +3.5 to +12.8 points. This gap shows that the reported scores reflect genuine infrared adaptation rather than residual visible-domain priors, and that \dataset{} is non-trivial: infrared alignment must be learned directly rather than inherited from RGB pretraining. To check one representative training run for seed sensitivity, we re-ran OpenAI CLIP ViT-B/32 fine-tuning under three random seeds with an otherwise identical recipe, obtaining a mean recall of $11.9 \pm 0.02$ (mean $\pm$ standard deviation over seeds). This check indicates low variance for that backbone, while the remaining backbone results are single-run measurements.

Table~\ref{tab:clip_persource} decomposes the best two backbones by source, retrieving within each source's own candidate pool. Difficulty varies sharply: SkyScript is hardest (mR$\approx$10.8), consistent with its noisy OpenStreetMap-derived supervision; RSICD and RSITMD show higher mR largely due to small pools (164 and 87 candidates), so per-source counts are reported. The aggregate in Table~\ref{tab:clip_retrieval} is dominated by the large, noisy RS5M and SkyScript pools.

Table~\ref{tab:clip_supervision_ablation} reports the supervision ablation motivating the final CLIP recipe on a development retrieval split, whose candidate pool is smaller and therefore not directly comparable to the 9,720-sample formal test split in Table~\ref{tab:clip_retrieval}. Averaged over five backbones, replacing original captions with IR-aware captions improves development IR-aware retrieval mean recall from 30.2\% to 43.0\%, showing that infrared-grounded text is a better language target for infrared retrieval under the ablation protocol. Original captions remain stronger on original-text retrieval because they preserve visible-domain wording, but that is not our target setting. A joint RGB--IR diagnostic obtains a similar IR-aware score but is not adopted, since our model-facing setting is infrared-modality inference. The ablation thus supports prioritizing IR-text alignment over RGB--IR matching.

\begin{table}[t]
  \centering
  \caption{CLIP supervision ablation on a smaller development retrieval split, averaged over five backbones. Original mR and IR-aware mR average R@1, R@5, and R@10 in both retrieval directions; these values are used for recipe comparison and are not directly comparable to the formal 9,720-sample test split.}
  \label{tab:clip_supervision_ablation}
  \scriptsize
  \setlength{\tabcolsep}{4.5pt}
  \renewcommand{\arraystretch}{1.0}
  \begin{tabular}{@{}lcc@{}}
    \toprule
    Supervision setting & Original mR & IR-aware mR \\
    \midrule
    Original-caption only & \textbf{35.1} & 30.2 \\
    IR-aware-caption only & 31.2 & \textbf{43.0} \\
    Joint RGB--IR diagnostic & 37.0 & 42.9 \\
    \bottomrule
  \end{tabular}
\end{table}

\subsection{IR-VLM Understanding Results}
\label{sec:vlm_results}
We deliberately use \emph{lexical proxy} diagnostics rather than a single accuracy score, because the corpus has no human-verified scene or object labels and the reference captions are machine-generated. Each metric is a keyword-match rate over the generated text, interpreted as a behavior signal, not semantic correctness: \textbf{IR-cue rate} is the fraction of answers with an infrared term (e.g.\ \emph{infrared, thermal, grayscale, intensity, contrast, texture}); \textbf{RGB-color rate} (lower better) the fraction with a visible-color word (e.g.\ \emph{green, blue, red, colorful}); \textbf{overclaim rate} (lower better) the fraction asserting unsupported physical detail (e.g.\ \emph{temperature, heat source, hot, cold}); \textbf{class-hit rate} the fraction containing a source-class token; and \textbf{average words} the response length. Thus a high IR-cue rate indicates infrared-style phrasing rather than verified grounding, and a low RGB-color rate indicates reduced leakage rather than stronger recognition. The ablations follow the actual development path: IR-aware text supervision for CLIP, and caption-only, class-aware, and multitask instruction data for VLMs.

Table~\ref{tab:vlm_clean} summarizes the clean diagnostic evaluation for the final IR-VLM recipe. Rather than a single accuracy score, each task is checked with script-level diagnostics reflecting the infrared instruction goal: IR-cue mentions, color leakage, overclaim, class-hit, and response length. The table reports averages over the six final backbones for each task type.

The diagnostic results show that the captioning prompt reliably elicits infrared-aware language (100.0\% IR-cue rate, almost no overclaiming). Infrared-cue explanation and visual-evidence QA also reference grayscale or thermal evidence frequently, but their higher overclaim rates show that longer explanatory answers introduce unsupported detail. Object-presence prompts are intentionally short, hence low IR-cue rate but nonzero class-hit. Class-hit stays near 1\% by design, as it measures literal repetition of the source class token (rare in infrared scene descriptions) and thus acts as a conservative lower-bound signal, not a recognition score. These patterns justify task-separated diagnostics: low RGB-color leakage alone does not imply strong grounding, and high infrared-cue frequency does not guarantee semantic precision.

\begin{table}[t]
  \centering
  \caption{Clean IR-VLM diagnostic evaluation on held-out infrared prompts, averaged over the six final VLM backbones. IR-cue rate, RGB-color rate, overclaim rate, and class-hit rate are percentages; average words measures response length.}
  \label{tab:vlm_clean}
  \scriptsize
  \setlength{\tabcolsep}{3.0pt}
  \renewcommand{\arraystretch}{1.0}
  \resizebox{\textwidth}{!}{%
  \begin{tabular}{@{}lccccc@{}}
    \toprule
    Task & IR-cue rate $\uparrow$ & RGB-color rate $\downarrow$ & Overclaim rate $\downarrow$ & Class-hit rate $\uparrow$ & Avg. words \\
    \midrule
    Infrared image captioning & \textbf{100.0} & \textbf{0.0} & 0.07 & \textbf{0.93} & 17.80 \\
    Scene question & 64.63 & \textbf{0.0} & 0.07 & 0.40 & 8.05 \\
    Object-presence question & 0.17 & \textbf{0.0} & \textbf{0.00} & 0.63 & 3.40 \\
    Infrared-cue explanation & 82.73 & 1.53 & 10.67 & 0.60 & 21.62 \\
    Visual-evidence QA & 63.63 & 0.07 & 19.87 & 0.83 & 22.20 \\
    \bottomrule
  \end{tabular}}
\end{table}

These patterns also differ across backbones under identical data and prompts. Some models produce longer, more explicitly infrared-aware answers, but this often comes with higher overclaiming because explanatory responses encourage unsupported thermal or physical detail. Others stay short and conservative, reducing hallucinated evidence but also offering weaker infrared-cue coverage. This backbone-dependent trade-off means no single model dominates every diagnostic, and it motivates reporting behavior-specific metrics rather than collapsing VLM performance into one aggregate score.

To confirm that the captioning behavior comes from IR-aware instruction tuning rather than the base models alone, Table~\ref{tab:vlm_zeroshot} compares zero-shot and fine-tuned captioning under the same prompts. The effect is large and consistent for captioning: fine-tuning raises the IR-cue rate to 100\% for every backbone (from as low as 4.0\% for LLaVA-1.5 and 16.2\% for H2RSVLM), eliminates visible-color leakage (e.g.\ H2RSVLM 16.0\%\,$\to$\,0.0\%, InstructBLIP 3.8\%\,$\to$\,0.0\%), and suppresses overclaiming (Qwen2.5-VL 28.8\%\,$\to$\,0.4\%). The gains are especially meaningful for models whose zero-shot responses either ignore infrared modality cues or revert to generic remote-sensing descriptions. This shows that IR-aware supervision, not the pretrained prior alone, is responsible for infrared-grounded captioning style, reduced color leakage, and more controlled use of thermal evidence.

\subsection{VLM Training-Stage Ablation}
\label{sec:vlm_stage_ablation}
The VLM training pipeline was developed in several stages, summarized in Table~\ref{tab:vlm_stage_ablation}. V1 used 5,000 caption-only samples, which were sufficient for teaching basic infrared description but too narrow for class-aware or evidence-seeking questions; the model learned to mention infrared cues, yet had little exposure to short discriminative answers or grounded reasoning prompts. V2 expanded to the full train split with class-aware scene hints, improving data coverage but weakening diagnostic balance because the supervision still lacked diverse instruction forms. The selected API-v3 multitask version combines 30,000 text-rewritten QA examples with 2,000 vision-audited examples in a conversation format, training all six backbones after removing malformed rows. This recipe preserves clean infrared-cue behavior while adding explicit scene, object, and evidence instruction types. It therefore gives the best balance of infrared-cue coverage, zero RGB-word leakage, and nonzero class-hit on the sanity set, while better matching the final evaluation tasks in Table~\ref{tab:vlm_clean}.

\begin{table}[t]
  \centering
  \caption{VLM training-stage ablation. Development metrics are averaged across the available backbones on a 12-case sanity set; they are used to select the training recipe, while Table~\ref{tab:vlm_clean} reports the formal clean-task evaluation.}
  \label{tab:vlm_stage_ablation}
  \scriptsize
  \setlength{\tabcolsep}{2.2pt}
  \renewcommand{\arraystretch}{1.0}
  \resizebox{\textwidth}{!}{%
  \begin{tabular}{@{}lclccc@{}}
    \toprule
    Stage & Data size & Main change & IR-cue rate & RGB-word rate & Dev class-hit \\
    \midrule
    Caption-only V1 & 5,000 & Single infrared caption prompt & 98.6 & 0.0 & -- \\
    Full class-aware V2 & 48,616 & Full train split with scene hints & 83.3 & 0.0 & 0.0 \\
    API-v3 multitask & 32,000 & 30K text QA + 2K vision-audited QA & \textbf{100.0} & \textbf{0.0} & \textbf{5.6} \\
    \bottomrule
  \end{tabular}}
\end{table}

\subsection{Dataset Quality Checks}
\label{sec:quality_checks}
The release-facing checks verify that the benchmark follows the intended split boundary: train--validation, train--test, and validation--test overlaps are all zero, and the referenced infrared paths have no missing-image cases. These checks are essential because \dataset{} draws from multiple collections with different naming rules, split conventions, and file organizations, where duplicate, stale, or mixed entries could silently inflate retrieval and instruction-tuning results. The RGB-leakage audit found 280 RGB-named paths in the original 10,000-image test split, which are excluded from the formal evaluation to leave the 9,720-image filtered split used in Table~\ref{tab:clip_retrieval}. To support reproducibility, error tracing, and independent auditing, the release package includes infrared images, IR-aware captions, split files, filtering logs, and evaluation scripts under a research-use setting; the source imagery inherits the licenses of its original collections. As the data derive from public remote-sensing collections and contain no personally identifiable information, we foresee no direct ethical concerns beyond the synthetic-infrared caveats, possible domain bias, and modality-transfer limitations noted above.
 
\begin{table}[t]
  \centering
  \caption{Zero-shot (original pretrained weights) vs.\ fine-tuned captioning diagnostics, showing the effect of IR-aware instruction tuning. Values are percentages.}
  \label{tab:vlm_zeroshot}
  \scriptsize
  \setlength{\tabcolsep}{3.0pt}
  \renewcommand{\arraystretch}{1.0}
  \resizebox{\textwidth}{!}{%
  \begin{tabular}{@{}lcccc@{}}
    \toprule
    Backbone & IR-cue (zero$\rightarrow$ft) & RGB-color (zero$\rightarrow$ft) & Overclaim (zero$\rightarrow$ft) & Avg.\ words (zero$\rightarrow$ft) \\
    \midrule
    Qwen2.5-VL-7B & 98.6 $\rightarrow$ 100.0 & 3.4 $\rightarrow$ 0.0 & 28.8 $\rightarrow$ 0.4 & 24.2 $\rightarrow$ 17.9 \\
    InstructBLIP-FLAN-T5-XL & 93.0 $\rightarrow$ 100.0 & 3.8 $\rightarrow$ 0.0 & 0.0 $\rightarrow$ 0.0 & 56.9 $\rightarrow$ 17.1 \\
    LLaVA-1.5-7B & 4.0 $\rightarrow$ 100.0 & 0.8 $\rightarrow$ 0.0 & 0.0 $\rightarrow$ 0.0 & 14.2 $\rightarrow$ 17.8 \\
    LLaVA-1.6-Vicuna-7B & 72.4 $\rightarrow$ 100.0 & 0.0 $\rightarrow$ 0.0 & 0.0 $\rightarrow$ 0.0 & 11.1 $\rightarrow$ 18.4 \\
    GeoChat-7B & 26.8 $\rightarrow$ 100.0 & 0.4 $\rightarrow$ 0.0 & 0.0 $\rightarrow$ 0.0 & 13.8 $\rightarrow$ 17.9 \\
    H2RSVLM-VHM-7B & 16.2 $\rightarrow$ 100.0 & 16.0 $\rightarrow$ 0.0 & 0.0 $\rightarrow$ 0.0 & 14.8 $\rightarrow$ 17.7 \\
    \bottomrule
  \end{tabular}}
\end{table}

\section{Conclusion}
\label{sec:conclusion}

This paper presents \dataset, a large-scale infrared remote-sensing vision-language dataset and benchmark connecting dataset construction, CLIP adaptation, VLM instruction tuning, and data-integrity evaluation under one infrared protocol. Beyond a set of infrared images, its contribution is an end-to-end workflow that rewrites visible-centric supervision into IR-aware text, enforces train-only hygiene, and evaluates models against zero-shot behavior where available. IR-aware text is essential: development retrieval mean recall rises from 30.2\% to 43.0\% over original captions under the ablation protocol, the strongest CLIP reaches 19.2\% ($+12.8$ over zero-shot) on the formal filtered test split, and fine-tuning drives VLM captioning IR-cue coverage to 100\% while reducing color leakage.

Two limitations bound the scope: the infrared modality is synthesized with DiffV2IR (closer to real infrared than grayscale, but not a sensor substitute), and IR-aware supervision is generated by Qwen2.5-VL-72B, so our diagnostics are lexical proxies rather than human-verified correctness. Future work can extend to real sensor data, more infrared bands, human-verified instructions, and full multi-seed evaluation.

\clearpage
\nocite{*}
\bibliographystyle{splncs04}
\bibliography{main}

\begin{thebibliography}{10}
\providecommand{\url}[1]{\texttt{#1}}
\providecommand{\urlprefix}{URL }
\providecommand{\doi}[1]{https://doi.org/#1}

\bibitem{astruc2025anysat}
Astruc, G., Gonthier, N., Mallet, C., Landrieu, L.: Anysat: One earth
  observation model for many resolutions, scales, and modalities. In:
  Proceedings of the Computer Vision and Pattern Recognition Conference. pp.
  19530--19540 (2025)

\bibitem{bai2025qwen25vltechnicalreport}
Bai, S., Chen, K., Liu, X., Wang, J., Ge, W., Song, S., Dang, K., Wang, P.,
  Wang, S., Tang, J., Zhong, H., Zhu, Y., Yang, M., Wan, J., Ding, W., Fu, Z.,
  Xu, Y., Ye, J., Zhang, X., Xie, T., Cheng, Z., Zhang, H., Yang, Z., Xu, H.,
  Lin, J.: Qwen2.5-vl technical report (2025)

\bibitem{bi2025llava}
Bi, J., Wang, Y., Chen, H., Xiao, X., Hecker, A., Tresp, V., Ma, Y.: Llava
  steering: Visual instruction tuning with 500x fewer parameters through
  modality linear representation-steering. In: Proceedings of the 63rd Annual
  Meeting of the Association for Computational Linguistics (Volume 1: Long
  Papers). pp. 15230--15250 (2025)

\bibitem{cao2026firemmir}
Cao, J., Liu, X., Xue, R.: Firemm-ir: An infrared-enhanced multi-modal large
  language model for comprehensive scene understanding in remote sensing forest
  fire monitoring. Sensors  \textbf{26}(2), ~390 (2026)

\bibitem{cao2024domain}
Cao, Q., Xu, Z., Chen, Y., Ma, C., Yang, X.: Domain prompt learning with
  quaternion networks. In: Proceedings of the IEEE/CVF Conference on Computer
  Vision and Pattern Recognition. pp. 26637--26646 (2024)

\bibitem{cao2025irgpt}
Cao, Z., Zhang, J., Zhang, R.: Irgpt: Understanding real-world infrared image
  with bi-cross-modal curriculum on large-scale benchmark. In: Proceedings of
  the IEEE/CVF International Conference on Computer Vision. pp. 166--176 (2025)

\bibitem{chen2025lrsclip}
Chen, W., Chen, J., Deng, Y., Chen, J., Feng, Y., Xi, Z., Liu, D., Li, K.,
  Meng, Y.: Lrsclip: A vision-language foundation model for aligning remote
  sensing image with longer text. In: Proceedings of the IEEE/CVF Conference on
  Computer Vision and Pattern Recognition (2025)

\bibitem{cheng2017remote}
Cheng, G., Han, J., Lu, X.: Remote sensing image scene classification:
  Benchmark and state of the art. Proceedings of the IEEE  \textbf{105}(10),
  1865--1883 (2017)

\bibitem{danish2025geobench}
Danish, M., Munir, M.A., Shah, S.R.A., Kuckreja, K., Khan, F.S., Fraccaro, P.,
  Lacoste, A., Khan, S.: Geobench-vlm: Benchmarking vision-language models for
  geospatial tasks. In: Proceedings of the IEEE/CVF International Conference on
  Computer Vision. pp. 7132--7142 (2025)

\bibitem{guo2024skysense}
Guo, X., Lao, J., Dang, B., Zhang, Y., Yu, L., Ru, L., Zhong, L., Huang, Z.,
  Wu, K., Hu, D., et~al.: Skysense: A multi-modal remote sensing foundation
  model towards universal interpretation for earth observation imagery. In:
  Proceedings of the IEEE/CVF Conference on Computer Vision and Pattern
  Recognition. pp. 27672--27683 (2024)

\bibitem{ha2017mfnet}
Ha, Q., Watanabe, K., Karasawa, T., Ushiku, Y., Harada, T.: Mfnet: Towards
  real-time semantic segmentation for autonomous vehicles with multi-spectral
  scenes. In: 2017 IEEE/RSJ International Conference on Intelligent Robots and
  Systems. pp. 5108--5115. IEEE (2017)

\bibitem{han2026fusionrs}
Han, J., Zhang, B., Sun, X., Zhang, Q., Dong, Y., Hu, C., Zhang, F., Wei, Y.,
  Guo, J.: Fusionrs: A large-scale rgb-infrared remote sensing dataset for
  dual-modal vision-language foundation models. arXiv preprint arXiv:2606.17020
   (2026)

\bibitem{han2023aviid}
Han, Z., Zhang, Z., Zhang, S., Zhang, G., Mei, S.: Aerial visible-to-infrared
  image translation: Dataset, evaluation, and baseline. Journal of Remote
  Sensing  \textbf{3}, ~0096 (2023)

\bibitem{jia2021llvip}
Jia, X., Zhu, C., Li, M., Tang, W., Zhou, W.: Llvip: A visible-infrared paired
  dataset for low-light vision. In: Proceedings of the IEEE/CVF International
  Conference on Computer Vision Workshops (ICCVW). pp. 3496--3504 (2021)

\bibitem{jiang2024infrared}
Jiang, S., Chen, Z., Liang, J., Zhao, Y., Liu, M., Qin, B.: Infrared-llava:
  Enhancing understanding of infrared images in multi-modal large language
  models. In: Findings of the Association for Computational Linguistics: EMNLP
  2024. pp. 8573--8591 (2024)

\bibitem{kuckreja2024geochat}
Kuckreja, K., Danish, M.S., Naseer, M., Das, A., Khan, S., Khan, F.S.: Geochat:
  Grounded large vision-language model for remote sensing. In: Proceedings of
  the IEEE/CVF conference on computer vision and pattern recognition. pp.
  27831--27840 (2024)

\bibitem{li2024s2mae}
Li, X., Hong, D., Chanussot, J.: S2mae: A spatial-spectral pretraining
  foundation model for spectral remote sensing data. In: Proceedings of the
  IEEE/CVF Conference on Computer Vision and Pattern Recognition. pp.
  24088--24097 (2024)

\bibitem{liu2024remoteclip}
Liu, F., Chen, D., Guan, Z., Zhou, X., Zhu, J., Ye, Q., Fu, L., Zhou, J.:
  Remoteclip: A vision language foundation model for remote sensing. IEEE
  Transactions on Geoscience and Remote Sensing  \textbf{62},  1--16 (2024)

\bibitem{liu2024improved}
Liu, H., Li, C., Li, Y., Lee, Y.J.: Improved baselines with visual instruction
  tuning. In: Proceedings of the IEEE/CVF conference on computer vision and
  pattern recognition. pp. 26296--26306 (2024)

\bibitem{liu2022m3fd}
Liu, J., Fan, X., Huang, Z., Wu, G., Liu, R., Zhong, W., Luo, Z.: Target-aware
  dual adversarial learning and a multi-scenario multi-modality benchmark to
  fuse infrared and visible for object detection. In: Proceedings of the
  IEEE/CVF Conference on Computer Vision and Pattern Recognition. pp.
  5802--5811 (2022)

\bibitem{lu2017rsicd}
Lu, X., Wang, B., Zheng, X., Li, X.: Exploring models and data for remote
  sensing image caption generation. IEEE Transactions on Geoscience and Remote
  Sensing  \textbf{56}(4),  2183--2195 (2017)

\bibitem{luo2025large}
Luo, J., Zhang, Y., Yang, X., Wu, K., Zhu, Q., Liang, L., Chen, J., Li, Y.:
  When large vision-language model meets large remote sensing imagery:
  Coarse-to-fine text-guided token pruning. In: Proceedings of the IEEE/CVF
  International Conference on Computer Vision. pp. 9206--9217 (2025)

\bibitem{mall2024remote}
Mall, U.K., Phoo, C.P., Liu, M., Vondrick, C., Hariharan, B., Bala, K.: Remote
  sensing vision-language foundation models without annotations via ground
  remote alignment. In: International Conference on Learning Representations.
  vol.~2024, pp. 49294--49314 (2024)

\bibitem{pang2025vhm}
Pang, C., Weng, X., Wu, J., Li, J., Liu, Y., Sun, J., Li, W., Wang, S., Feng,
  L., Xia, G.S., He, C.: Vhm: Versatile and honest vision language model for
  remote sensing image analysis. In: Proceedings of the AAAI Conference on
  Artificial Intelligence. vol.~39, pp. 6381--6388 (2025)

\bibitem{ran2025diffv2ir}
Ran, L., Wang, L., Wang, G., Wang, P., Zhang, Y.: Diffv2ir: Visible-to-infrared
  diffusion model via vision-language understanding. arXiv preprint
  arXiv:2503.19012  (2025)

\bibitem{razakarivony2016vedai}
Razakarivony, S., Jurie, F.: Vehicle detection in aerial imagery: A small
  target detection benchmark. Journal of Visual Communication and Image
  Representation  \textbf{34},  187--203 (2016)

\bibitem{safaei2025filter}
Safaei, B., Siddiqui, F., Xu, J., Patel, V.M., Lo, S.Y.: Filter images first,
  generate instructions later: Pre-instruction data selection for visual
  instruction tuning. In: Proceedings of the IEEE/CVF Conference on Computer
  Vision and Pattern Recognition. pp. 14247--14256 (2025)

\bibitem{soni2025earthdial}
Soni, S., Dudhane, A., Debary, H., Fiaz, M., Munir, M.A., Danish, M.S.,
  Fraccaro, P., Watson, C.D., Klein, L.J., Khan, F.S., et~al.: Earthdial:
  Turning multi-sensory earth observations to interactive dialogues. In:
  Proceedings of the Computer Vision and Pattern Recognition Conference. pp.
  14303--14313 (2025)

\bibitem{sun2022dronevehicle}
Sun, Y., Cao, B., Zhu, P., Hu, Q.: Drone-based rgb-infrared cross-modality
  vehicle detection via uncertainty-aware learning. IEEE Transactions on
  Circuits and Systems for Video Technology  \textbf{32}(10),  6700--6713
  (2022)

\bibitem{wang2024skyscript}
Wang, Z., Prabha, R., Huang, T., Wu, J., Rajagopal, R.: Skyscript: A large and
  semantically diverse vision-language dataset for remote sensing. In:
  Proceedings of the AAAI Conference on Artificial Intelligence. vol.~38, pp.
  5805--5813 (2024)

\bibitem{wen2024makes}
Wen, X., Zhao, B., Chen, Y., Pang, J., Qi, X.: What makes clip more robust to
  long-tailed pre-training data? a controlled study for transferable insights.
  Advances in Neural Information Processing Systems  \textbf{37},  36567--36601
  (2024)

\bibitem{yuan2021rsitmd}
Yuan, Z., Zhang, W., Fu, K., Li, X., Deng, C., Wang, H., Sun, X.: Exploring a
  fine-grained multiscale method for cross-modal remote sensing image
  retrieval. IEEE Transactions on Geoscience and Remote Sensing  \textbf{60},
  1--19 (2021)

\bibitem{zanella2024boosting}
Zanella, M., G{\'e}rin, B., Ayed, I.B.: Boosting vision-language models with
  transduction. Advances in Neural Information Processing Systems  \textbf{37},
   62223--62256 (2024)

\bibitem{zhang2025llava}
Zhang, S., Fang, Q., Yang, Y., Feng, Y.: Llava-mini: Efficient image and video
  large multimodal models with one vision token. In: International Conference
  on Learning Representations. vol.~2025, pp. 53285--53310 (2025)

\bibitem{zhang2024rs5m}
Zhang, Z., Zhao, T., Guo, Y., Yin, J.: Rs5m and georsclip: A large-scale
  vision-language dataset and a large vision-language model for remote sensing.
  IEEE Transactions on Geoscience and Remote Sensing  \textbf{62},  1--23
  (2024)

\bibitem{zhu2025skysense}
Zhu, Q., Lao, J., Ji, D., Luo, J., Wu, K., Zhang, Y., Ru, L., Wang, J., Chen,
  J., Yang, M., et~al.: Skysense-o: Towards open-world remote sensing
  interpretation with vision-centric visual-language modeling. In: Proceedings
  of the IEEE/CVF Conference on Computer Vision and Pattern Recognition. pp.
  14733--14744 (2025)

\end{thebibliography}

\clearpage
\appendix
\renewcommand{\theHsection}{appendix.\Alph{section}}
\renewcommand{\theHsubsection}{appendix.\Alph{section}.\arabic{subsection}}
\renewcommand{\theHfigure}{appendix.\arabic{figure}}
\renewcommand{\theHtable}{appendix.\arabic{table}}
\renewcommand{\theHalgorithm}{appendix.\arabic{algorithm}}
\section{Mathematical Formulation and Algorithms}
\label{app:formulation_algorithms}

This appendix formalizes the dataset construction, model adaptation, and evaluation metrics used in the main paper. The notation follows the implementation used for the reported \irclip{} and \irvlm{} experiments: CLIP models are fine-tuned with a symmetric image-text contrastive loss, VLMs are adapted with instruction-following loss and LoRA adapters, and VLM behavior is evaluated with lexical diagnostics rather than human-verified semantic accuracy.

\subsection{Notation}
\label{app:notation}

Let the source remote-sensing pool be
\begin{equation}
  \mathcal{S}
  =
  \{(I_i^{\mathrm{rgb}}, c_i, s_i, d_i)\}_{i=1}^{N},
  \qquad N=600{,}000,
\end{equation}
where $I_i^{\mathrm{rgb}}$ is the visible-band source image, $c_i$ is the original source caption or metadata text, $s_i \in \{\mathrm{train},\mathrm{val},\mathrm{test}\}$ is the split, and $d_i$ is the source dataset identifier. The model-facing infrared image is generated by a visible-to-infrared translation function,
\begin{equation}
  I_i^{\mathrm{ir}} = T_{\phi}(I_i^{\mathrm{rgb}}),
\end{equation}
where $T_{\phi}$ denotes the DiffV2IR-based synthesis pipeline used in the paper. The IR-aware text is written as
\begin{equation}
  t_i = R_{\omega}(I_i^{\mathrm{rgb}}, I_i^{\mathrm{ir}}, c_i),
\end{equation}
where $R_{\omega}$ denotes the IR-aware caption rewriting process. The RGB image is used only at construction time to preserve scene semantics during rewriting; the model-facing example used for fine-tuning and evaluation is
\begin{equation}
  x_i = (I_i^{\mathrm{ir}}, t_i, s_i, d_i).
\end{equation}

For CLIP-style fine-tuning, each retained training item is an image-text pair $(I_i^{\mathrm{ir}}, t_i)$. For VLM instruction tuning, a train-only subset of retained records is converted into an instruction triple
\begin{equation}
  z_i = (I_i^{\mathrm{ir}}, q_i, a_i),
\end{equation}
where $q_i$ is a task prompt and $a_i$ is the target answer. The final API-v3 multitask VLM recipe uses 32,000 train-split conversations constructed from the retained training pool.

\subsection{Filtering and Split Hygiene}
\label{app:filtering}

The retained IR-aware caption set is defined by a binary validity predicate
\begin{equation}
  m_i
  =
  \mathbf{1}
  \left[
  I_i^{\mathrm{ir}}\ \mathrm{exists}
  \ \wedge\
  |t_i| > 0
  \ \wedge\
  s_i \in \{\mathrm{train}, \mathrm{val}, \mathrm{test}\}
  \ \wedge\
  Q(x_i)=\mathrm{pass}
  \right],
\end{equation}
where $Q(\cdot)$ summarizes the quality checks for empty entries, malformed split fields, missing images, weak IR descriptions, and unsupported text. The retained split-specific sets are
\begin{equation}
  \mathcal{D}_{s}
  =
  \{x_i : m_i=1,\ s_i=s\},
  \qquad
  s \in \{\mathrm{train},\mathrm{val},\mathrm{test}\}.
\end{equation}
The paper uses $|\mathcal{D}_{\mathrm{train}}|=48{,}616$, $|\mathcal{D}_{\mathrm{val}}|=416$, and $|\mathcal{D}_{\mathrm{test}}|=10{,}000$ IR-aware caption records. CLIP uses the retained train records as image-text pairs. The final VLM recipe uses $\mathcal{Z}_{\mathrm{train}}\subset\mathcal{D}_{\mathrm{train}}$, with $|\mathcal{Z}_{\mathrm{train}}|=32{,}000$ conversations.

For the formal CLIP retrieval evaluation, we further apply a path-level audit and remove test entries whose path strings indicate residual visible-band naming. Let
\begin{equation}
  \rho_i = \mathbf{1}[\mathrm{``rgb''}\in \mathrm{lowercase}(\mathrm{path}(I_i^{\mathrm{ir}}))].
\end{equation}
The filtered evaluation set is
\begin{equation}
  \mathcal{D}_{\mathrm{test}}^{\mathrm{clean}}
  =
  \{x_i \in \mathcal{D}_{\mathrm{test}} : \rho_i=0\},
  \qquad
  |\mathcal{D}_{\mathrm{test}}^{\mathrm{clean}}|=9{,}720.
\end{equation}
This audit is deliberately conservative and file-name based: it prevents path-level visible-band leakage in the reported retrieval surface, but it is not a content-level detector of all possible source-domain artifacts. The release keeps the flagged list so that stronger content-level audits can be added without changing the reported split definition.
Equivalently, the model-facing sets used by the two training stages are
\begin{align}
  \mathcal{P}_{\mathrm{CLIP}}
  &=
  \{(I_i^{\mathrm{ir}},t_i):x_i\in\mathcal{D}_{\mathrm{train}}\},
  \\
  \mathcal{Z}_{\mathrm{train}}
  &=
  \{G_{\xi}(x_i):x_i\in\mathcal{D}_{\mathrm{train}},\ h_{\xi}(x_i)=1\},
  \qquad
  |\mathcal{Z}_{\mathrm{train}}|=32{,}000,
  \\
  \mathcal{E}_{\mathrm{CLIP}}
  &=
  \mathcal{D}_{\mathrm{test}}^{\mathrm{clean}},
\end{align}
where $G_{\xi}$ converts a retained train record into an instruction tuple and $h_{\xi}$ denotes the API-v3 multitask selection and validity filter.

\subsection{Synthetic-IR Realism Metrics}
\label{app:ir_realism_metrics}

To compare synthetic infrared images with real infrared images from AVIID, the main paper reports FID and grayscale-histogram distances. Let $\mathcal{X}$ and $\mathcal{Y}$ be two image sets, and let $\mu_{\mathcal{X}},\Sigma_{\mathcal{X}}$ and $\mu_{\mathcal{Y}},\Sigma_{\mathcal{Y}}$ be the mean and covariance of their Inception features. The Fréchet Inception Distance is
\begin{equation}
  \mathrm{FID}(\mathcal{X},\mathcal{Y})
  =
  \|\mu_{\mathcal{X}}-\mu_{\mathcal{Y}}\|_2^2
  +
  \mathrm{Tr}
  \left(
  \Sigma_{\mathcal{X}}+\Sigma_{\mathcal{Y}}
  -
  2(\Sigma_{\mathcal{X}}\Sigma_{\mathcal{Y}})^{1/2}
  \right).
\end{equation}

For grayscale histograms, let $p,q \in \mathbb{R}^{B}$ be normalized intensity histograms with $\sum_{b=1}^{B} p_b=\sum_{b=1}^{B} q_b=1$. With a small numerical constant $\epsilon>0$, the histogram metrics are
\begin{align}
  D_{\mathrm{KL}}(p\|q)
  &=
  \sum_{b=1}^{B} p_b
  \log\frac{p_b+\epsilon}{q_b+\epsilon},
  \\
  D_{\mathrm{JS}}(p,q)
  &=
  \frac{1}{2}D_{\mathrm{KL}}(p\|m)
  +
  \frac{1}{2}D_{\mathrm{KL}}(q\|m),
  \qquad
  m=\frac{p+q}{2},
  \\
  \chi^2(p,q)
  &=
  \frac{1}{2}
  \sum_{b=1}^{B}
  \frac{(p_b-q_b)^2}{p_b+q_b+\epsilon},
  \\
  \mathrm{HI}(p,q)
  &=
  \sum_{b=1}^{B}\min(p_b,q_b).
\end{align}
Lower values are better for FID, KL, JS, and $\chi^2$, while higher values are better for histogram intersection.

\subsection{\irclip{} Contrastive Objective}
\label{app:clip_objective}

For a mini-batch of $B$ retained image-text pairs $\{(I_i^{\mathrm{ir}},t_i)\}_{i=1}^{B}$, the image encoder $f_{\theta}$ and text encoder $g_{\psi}$ produce normalized embeddings
\begin{equation}
  u_i =
  \frac{f_{\theta}(I_i^{\mathrm{ir}})}
       {\|f_{\theta}(I_i^{\mathrm{ir}})\|_2},
  \qquad
  v_i =
  \frac{g_{\psi}(t_i)}
       {\|g_{\psi}(t_i)\|_2}.
\end{equation}
The implementation freezes the text side when \texttt{freeze-text} is enabled and updates the visual tower, projection parameters, and logit scale. Let $\alpha=\exp(\gamma)$ be the learned logit scale. The image-text similarity logits are
\begin{equation}
  z_{ij} = \alpha\, u_i^{\top}v_j.
\end{equation}
The image-to-text and text-to-image losses are
\begin{align}
  \mathcal{L}_{I\rightarrow T}
  &=
  -\frac{1}{B}
  \sum_{i=1}^{B}
  \log
  \frac{\exp(z_{ii})}
       {\sum_{j=1}^{B}\exp(z_{ij})},
  \\
  \mathcal{L}_{T\rightarrow I}
  &=
  -\frac{1}{B}
  \sum_{i=1}^{B}
  \log
  \frac{\exp(z_{ii})}
       {\sum_{j=1}^{B}\exp(z_{ji})}.
\end{align}
The final symmetric contrastive loss is
\begin{equation}
  \mathcal{L}_{\mathrm{CLIP}}
  =
  \frac{1}{2}
  \left(
  \mathcal{L}_{I\rightarrow T}
  +
  \mathcal{L}_{T\rightarrow I}
  \right).
\end{equation}

\subsection{Retrieval Metrics}
\label{app:retrieval_metrics}

For a retrieval set of $M$ paired examples, let $U$ and $V$ denote the normalized image and text embeddings. The correct text for image $i$ is $v_i$, and the correct image for text $i$ is $u_i$. The rank of the paired text under image-to-text retrieval is
\begin{equation}
  r_i^{I\rightarrow T}
  =
  1+
  \sum_{j\neq i}
  \mathbf{1}
  [u_i^{\top}v_j > u_i^{\top}v_i].
\end{equation}
The rank $r_i^{T\rightarrow I}$ is defined analogously by ranking images for text query $v_i$. Recall at $K$ is
\begin{equation}
  R_{I\rightarrow T}@K
  =
  \frac{1}{M}
  \sum_{i=1}^{M}
  \mathbf{1}[r_i^{I\rightarrow T}\le K],
\end{equation}
with $R_{T\rightarrow I}@K$ defined similarly. The mean recall reported in the paper averages the three recall cutoffs in both retrieval directions:
\begin{equation}
  \mathrm{mR}
  =
  \frac{1}{6}
  \sum_{K\in\{1,5,10\}}
  \left(
  R_{I\rightarrow T}@K
  +
  R_{T\rightarrow I}@K
  \right).
\end{equation}
The same definition is used for the synthetic-IR retrieval benchmark and the AVIID paired retrieval sanity check, with only the query and gallery modalities changed.
Checkpoint selection is written as
\begin{align}
  c^{\star}
  &=
  \arg\max_{c\in\mathcal{C}}
  \mathrm{mR}_{\mathrm{val}}(c),
  \\
  \mathrm{Score}_{\mathrm{test}}
  &=
  \{R_{I\rightarrow T}@K,\ R_{T\rightarrow I}@K\}_{K\in\{1,5,10\}}
  \big|_{\mathcal{D}_{\mathrm{test}}^{\mathrm{clean}},c^{\star}}.
\end{align}

\subsection{\irvlm{} Instruction-Tuning Objective}
\label{app:vlm_objective}

For VLM training, each example is represented as $(I_i^{\mathrm{ir}},q_i,a_i)\in\mathcal{Z}_{\mathrm{train}}$, where $q_i$ is the instruction and $a_i=(a_{i,1},\ldots,a_{i,L_i})$ is the target answer. The visual encoder $E_{\eta}$ produces visual tokens, and a multimodal projector $P_{\omega}$ maps them into the language-model embedding space when the backbone exposes such a projector:
\begin{equation}
  H_i = P_{\omega}(E_{\eta}(I_i^{\mathrm{ir}})).
\end{equation}
The instruction-tuned language model predicts answer tokens conditioned on the visual tokens and the user prompt:
\begin{equation}
  p_{\Theta}(a_i \mid I_i^{\mathrm{ir}},q_i)
  =
  \prod_{\ell=1}^{L_i}
  p_{\Theta}
  (a_{i,\ell}\mid H_i,q_i,a_{i,<\ell}).
\end{equation}
Only assistant-answer tokens contribute to the supervised loss; prompt tokens are masked. The training objective is
\begin{equation}
  \mathcal{L}_{\mathrm{VLM}}
  =
  -\frac{1}{\sum_i L_i}
  \sum_{i}
  \sum_{\ell=1}^{L_i}
  \log
  p_{\Theta}
  (a_{i,\ell}\mid H_i,q_i,a_{i,<\ell}).
\end{equation}
The selected adapter parameters are therefore
\begin{equation}
  \Theta^{\star}
  =
  \arg\min_{\Theta_{\mathrm{LoRA}},\,\omega}
  \mathcal{L}_{\mathrm{VLM}}(\mathcal{Z}_{\mathrm{train}};\Theta,\omega),
\end{equation}
with the base language model parameters fixed.

Parameter-efficient adaptation is implemented with LoRA. For a frozen linear layer $W\in\mathbb{R}^{d_{\mathrm{out}}\times d_{\mathrm{in}}}$, LoRA replaces the effective weight by
\begin{equation}
  W'
  =
  W
  +
  \Delta W,
  \qquad
  \Delta W
  =
  \frac{\alpha_{\mathrm{LoRA}}}{r}BA,
\end{equation}
where $A\in\mathbb{R}^{r\times d_{\mathrm{in}}}$ and $B\in\mathbb{R}^{d_{\mathrm{out}}\times r}$ are trainable low-rank matrices. The experiments use $r=32$, $\alpha_{\mathrm{LoRA}}=64$, and dropout $0.05$. For Qwen2.5-VL, LoRA is applied to attention and MLP projection modules. For InstructBLIP, LoRA is applied to the query and value modules. For LLaVA-style, GeoChat, and H2RSVLM-style models, LoRA adapters and the exposed multimodal projector are trained while the base LLM is kept fixed.

\subsection{VLM Diagnostic Metrics}
\label{app:vlm_metrics}

The VLM metrics in the main paper are lexical diagnostics, not human-verified semantic accuracy. For $n$ generated answers $\{y_i\}_{i=1}^{n}$, define keyword sets $\mathcal{K}_{\mathrm{IR}}$, $\mathcal{K}_{\mathrm{RGB}}$, and $\mathcal{K}_{\mathrm{over}}$ for infrared evidence terms, visible-color terms, and unsupported thermal overclaim terms. Let $\mathrm{hit}(y,\mathcal{K})$ be one when any keyword in $\mathcal{K}$ appears in $y$, and zero otherwise. Then
\begin{align}
  \mathrm{IRCueRate}
  &=
  \frac{100}{n}
  \sum_{i=1}^{n}
  \mathrm{hit}(y_i,\mathcal{K}_{\mathrm{IR}}),
  \\
  \mathrm{RGBColorRate}
  &=
  \frac{100}{n}
  \sum_{i=1}^{n}
  \mathrm{hit}(y_i,\mathcal{K}_{\mathrm{RGB}}),
  \\
  \mathrm{OverclaimRate}
  &=
  \frac{100}{n}
  \sum_{i=1}^{n}
  \mathrm{hit}(y_i,\mathcal{K}_{\mathrm{over}}).
\end{align}
The class-hit diagnostic uses a source class token set $\mathcal{C}_i$ extracted from the source class name or task metadata:
\begin{equation}
  \mathrm{ClassHitRate}
  =
  \frac{100}{n}
  \sum_{i=1}^{n}
  \mathbf{1}
  [\mathrm{tokens}(y_i)\cap \mathcal{C}_i \neq \emptyset].
\end{equation}
The average response length is
\begin{equation}
  \mathrm{AvgWords}
  =
  \frac{1}{n}
  \sum_{i=1}^{n}
  |\mathrm{words}(y_i)|.
\end{equation}
The reported task-level diagnostic vector is
\begin{equation}
  \mathbf{m}_{\tau}
  =
  \begin{bmatrix}
  \mathrm{IRCueRate}_{\tau} \\
  \mathrm{RGBColorRate}_{\tau} \\
  \mathrm{OverclaimRate}_{\tau} \\
  \mathrm{ClassHitRate}_{\tau} \\
  \mathrm{AvgWords}_{\tau}
  \end{bmatrix},
\end{equation}
where $\tau$ indexes the prompt type.
These diagnostics are intentionally conservative. A high IR-cue rate indicates that the response uses infrared-style evidence language, while a low RGB-color rate indicates reduced visible-color leakage. Neither should be interpreted as object-level or scene-level semantic accuracy.

Algorithms~\ref{alg:dataset_construction}--\ref{alg:vlm_training} summarize how the equations above are instantiated in the experimental pipeline. Each algorithm keeps the operational step in text and places the actual transformation or selection rule in mathematical form, so that the appendix remains reproducible without duplicating implementation details from the code.

\begin{algorithm}[t]
\caption{Infrared record construction}
\label{alg:dataset_construction}
\begin{algorithmic}[1]
\Require $\mathcal{S},T_{\phi},R_{\omega},Q,h_{\xi},G_{\xi}$
\Ensure $\{\mathcal{D}_s\}_{s}$, $\mathcal{P}_{\mathrm{CLIP}}$, $\mathcal{Z}_{\mathrm{train}}$, $\mathcal{D}_{\mathrm{test}}^{\mathrm{clean}}$
\For{$i=1,\ldots,N$}
  \State \textbf{Synthesize and rewrite:} $I_i^{\mathrm{ir}}\gets T_{\phi}(I_i^{\mathrm{rgb}})$,\quad $t_i\gets R_{\omega}(I_i^{\mathrm{rgb}},I_i^{\mathrm{ir}},c_i)$
  \State \textbf{Form one candidate record:} $x_i\gets(I_i^{\mathrm{ir}},t_i,s_i,d_i)$
  \State \textbf{Apply quality gate:} $m_i\gets \mathbf{1}[I_i^{\mathrm{ir}}\ \mathrm{exists}\wedge |t_i|>0$
  \Statex \hspace{\algorithmicindent}$\wedge\ s_i\in\{\mathrm{train},\mathrm{val},\mathrm{test}\}\wedge Q(x_i)=\mathrm{pass}]$
\EndFor
\State \textbf{Retain split records:} $\mathcal{D}_s\gets\{x_i:m_i=1,\ s_i=s\}$,\quad $s\in\{\mathrm{train},\mathrm{val},\mathrm{test}\}$
\State \textbf{Derive CLIP pairs:} $\mathcal{P}_{\mathrm{CLIP}}\gets\{(I_i^{\mathrm{ir}},t_i):x_i\in\mathcal{D}_{\mathrm{train}}\}$
\State \textbf{Derive VLM conversations:} $\mathcal{Z}_{\mathrm{train}}\gets\{G_{\xi}(x_i):x_i\in\mathcal{D}_{\mathrm{train}},\ h_{\xi}(x_i)=1\}$
\State \textbf{Clean retrieval test:} $\mathcal{D}_{\mathrm{test}}^{\mathrm{clean}}\gets\{x_i\in\mathcal{D}_{\mathrm{test}}:\rho_i=0\}$
\end{algorithmic}
\end{algorithm}

\begin{algorithm}[t]
\caption{\irclip{} contrastive adaptation and retrieval}
\label{alg:clip_training}
\begin{algorithmic}[1]
\Require $\mathcal{P}_{\mathrm{CLIP}}$, $\mathcal{D}_{\mathrm{val}}$, $\mathcal{D}_{\mathrm{test}}^{\mathrm{clean}}$
\Ensure $c^{\star}$, $\mathrm{Score}_{\mathrm{test}}$
\State \textbf{Set trainable parameters:} $\Omega_{\mathrm{CLIP}}=\{\theta_{\mathrm{vis}},\theta_{\mathrm{proj}},\gamma\}$,\quad $\psi\gets\psi_{0}$
\For{$e=1,\ldots,E$}
  \State \textbf{Optimize contrastive objective:}
  \Statex \hspace{\algorithmicindent}$\Omega_{\mathrm{CLIP}}^{(e)}\gets\arg\min_{\Omega_{\mathrm{CLIP}}}\sum_{(I,t)\in\mathcal{P}_{\mathrm{CLIP}}}
  \mathcal{L}_{\mathrm{CLIP}}(I,t;\Omega_{\mathrm{CLIP}},\psi_{0})$
  \State \textbf{Save candidate checkpoint:} $c_e\gets\Omega_{\mathrm{CLIP}}^{(e)}$
\EndFor
\State \textbf{Select by validation retrieval:} $c^{\star}\gets\arg\max_{c\in\{c_e\}_{e=1}^{E}}\mathrm{mR}_{\mathrm{val}}(c)$
\State \textbf{Report held-out retrieval:}
\Statex \hspace{\algorithmicindent}$\mathrm{Score}_{\mathrm{test}}\gets\{R_{I\rightarrow T}@K,R_{T\rightarrow I}@K\}_{K\in\{1,5,10\}}\big|_{\mathcal{D}_{\mathrm{test}}^{\mathrm{clean}},c^{\star}}$
\end{algorithmic}
\end{algorithm}

\begin{algorithm}[t]
\caption{\irvlm{} instruction adaptation and diagnostics}
\label{alg:vlm_training}
\begin{algorithmic}[1]
\Require $\mathcal{Z}_{\mathrm{train}}$, $\mathcal{H}_{\tau}$, $\Theta_{0}$
\Ensure $\Theta^{\star}$ and $\{\mathbf{m}_{\tau}\}_{\tau}$
\State \textbf{Set trainable parameters:} $\Omega_{\mathrm{VLM}}=\{\Theta_{\mathrm{LoRA}},\omega\}$,\quad $\Theta_{\mathrm{base}}\gets\Theta_{0}$
\State \textbf{Fit adapters on train-only conversations:}
\Statex \hspace{\algorithmicindent}$\Omega_{\mathrm{VLM}}^{\star}\gets
\arg\min_{\Omega_{\mathrm{VLM}}}\mathcal{L}_{\mathrm{VLM}}(\mathcal{Z}_{\mathrm{train}};\Omega_{\mathrm{VLM}},\Theta_{0})$
\ForAll{$\tau$}
  \ForAll{$(I_j^{\mathrm{ir}},q_j)\in\mathcal{H}_{\tau}$}
    \State \textbf{Generate answer:} $y_j\gets\arg\max_y p_{\Omega_{\mathrm{VLM}}^{\star},\Theta_{0}}(y\mid I_j^{\mathrm{ir}},q_j)$
  \EndFor
  \State \textbf{Compute lexical diagnostics:} $\mathbf{m}_{\tau}\gets[\mathrm{IR}_{\tau},\mathrm{RGB}_{\tau},\mathrm{Over}_{\tau},\mathrm{Class}_{\tau},\mathrm{Len}_{\tau}]$
\EndFor
\State $\Theta^{\star}\gets(\Omega_{\mathrm{VLM}}^{\star},\Theta_{0})$
\end{algorithmic}
\end{algorithm}

\subsection{Implementation Traceability}
\label{app:traceability}

The equations above correspond to the experimental scripts used to produce the paper tables. The CLIP objective follows the symmetric cross-entropy implementation used for both HuggingFace CLIP and OpenCLIP backbones. Retrieval metrics follow the evaluation routine that computes top-$K$ hits from the image-text similarity matrix and then averages $R@1$, $R@5$, and $R@10$ across the two directions. The VLM objective follows the train-only multitask QA scripts, where labels are applied only to assistant answer tokens. The VLM diagnostic metrics follow the clean evaluation scripts that count infrared evidence words, visible-color words, overclaim words, class-token hits, and response length.

\clearpage
\section{Supplementary Experimental Results}
\label{app:supp_results}

This appendix expands the aggregate tables of the main paper into per-backbone, per-source, and per-configuration form. It is intended to expose the variation that is compressed by averaged results, including backbone-specific retrieval behavior, source-dependent gallery difficulty, auxiliary paired-retrieval transfer, and seed-level stability. All numbers are taken directly from the evaluation outputs used for the main paper; no values are re-estimated or post-processed beyond table formatting.

\subsection{Extended Per-Source Retrieval}
\label{app:per_source_retrieval}

Table~\ref{tab:per_source_full} gives the full bidirectional recall behind the per-source summary of the main paper, for the two strongest CLIP backbones, retrieving within each source's own candidate pool. This breakdown is useful because a single pooled score mixes backbone alignment quality, gallery size, caption noise, and repeated remote-sensing layouts. The source-difficulty ordering is consistent across both backbones: SkyScript is hardest because its OpenStreetMap-derived supervision is noisier and less caption-like, while the small RSICD and RSITMD pools inflate recall by presenting fewer competing candidates. We therefore report their sample counts directly. Per-source retrieval was computed only for these two backbones; for completeness, Table~\ref{tab:pooled_mr} lists the pooled mean recall of all five CLIP backbones on the merged 9{,}720-image split.

\vspace{-2.0em}
\begin{table}[H]
  \centering
  \caption{Full per-source bidirectional recall (\%) for the two strongest CLIP backbones on the filtered infrared test split, retrieving within each source's candidate pool. I2T is image-to-text, T2I is text-to-image; mR averages all six recall values.}
  \label{tab:per_source_full}
  \scriptsize
  \setlength{\tabcolsep}{4pt}
  \renewcommand{\arraystretch}{1.0}
  \resizebox{\textwidth}{!}{%
  \begin{tabular}{@{}ccccccccc@{}}
    \toprule
    Source & \# & I2T R@1 & I2T R@5 & I2T R@10 & T2I R@1 & T2I R@5 & T2I R@10 & mR \\
    \midrule
    \multicolumn{9}{@{}c@{}}{\cellcolor{oursrowgreen}\textbf{OpenAI CLIP ViT-L/14}} \\
    RS5M & 7,858 & 10.13 & 23.61 & 32.09 & 11.53 & 25.49 & 33.76 & 22.77 \\
    SkyScript & 1,096 & 3.28 & 9.31 & 15.51 & 4.47 & 12.77 & 19.43 & 10.80 \\
    NWPU & 515 & 12.23 & 34.56 & 49.32 & 11.84 & 40.58 & 55.15 & 33.95 \\
    RSICD & 164 & 15.24 & 40.85 & 54.27 & 13.41 & 43.29 & 59.15 & 37.70 \\
    RSITMD & 87 & 17.24 & 57.47 & 77.01 & 28.74 & 66.67 & 80.46 & 54.60 \\
    \midrule
    \multicolumn{9}{@{}c@{}}{\cellcolor{oursrowgreen}\textbf{GeoRSCLIP ViT-B/32}} \\
    RS5M & 7,858 & 6.58 & 16.80 & 24.13 & 6.62 & 16.79 & 24.14 & 15.84 \\
    SkyScript & 1,096 & 3.56 & 9.31 & 14.96 & 4.38 & 13.23 & 19.07 & 10.75 \\
    NWPU & 515 & 12.04 & 32.23 & 48.74 & 12.23 & 34.17 & 50.68 & 31.68 \\
    RSICD & 164 & 12.20 & 32.32 & 48.17 & 13.41 & 36.59 & 53.05 & 32.62 \\
    RSITMD & 87 & 22.99 & 60.92 & 74.71 & 25.29 & 64.37 & 79.31 & 54.60 \\
    \bottomrule
  \end{tabular}}
\end{table}
\vspace{-2.0em}
\vspace{-1.0em}

The per-source table explains why the same checkpoint can look stronger or weaker depending on the gallery being queried. Smaller caption datasets produce higher recall partly because they have fewer competing candidates, while larger sources expose noisier supervision and more visually repeated layouts. This is why the appendix reports both source-specific and pooled retrieval instead of using only the most favorable view.

\begin{table}[H]
  \centering
  \caption{Pooled bidirectional recall (\%) of all five CLIP backbones on the merged 9{,}720-image filtered test split (all sources in one candidate pool). mR averages the six recall values and matches the fine-tuned column of the main-paper retrieval table.}
  \label{tab:pooled_mr}
  \scriptsize
  \setlength{\tabcolsep}{4pt}
  \renewcommand{\arraystretch}{1.0}
  \resizebox{\textwidth}{!}{%
  \begin{tabular}{@{}cccccccc@{}}
    \toprule
    Backbone & I2T R@1 & I2T R@5 & I2T R@10 & T2I R@1 & T2I R@5 & T2I R@10 & mR \\
    \midrule
    OpenAI CLIP ViT-L/14 & 8.33 & 19.78 & 27.14 & 9.50 & 21.43 & 28.80 & \textbf{19.16} \\
    GeoRSCLIP ViT-B/32 & 5.32 & 14.15 & 20.43 & 5.56 & 14.22 & 20.57 & 13.37 \\
    OpenCLIP ViT-B/32 & 5.13 & 13.01 & 18.48 & 6.21 & 14.15 & 19.81 & 12.80 \\
    OpenAI CLIP ViT-B/32 & 4.76 & 12.46 & 17.69 & 5.49 & 13.05 & 18.52 & 11.99 \\
    RemoteCLIP ViT-B/32 & 1.14 & 3.97 & 6.50 & 1.48 & 4.72 & 7.82 & 4.27 \\
    \bottomrule
  \end{tabular}}
\end{table}

The pooled results in Table~\ref{tab:pooled_mr} should be read together with Table~\ref{tab:per_source_full}. Per-source evaluation explains where the model succeeds or struggles, whereas pooled evaluation reflects the harder deployment-like setting in which all sources compete in the same gallery. The gap between RS5M/SkyScript and the smaller caption datasets also warns against over-reading per-source recall without considering candidate-pool size and supervision noise.

\subsection{Auxiliary AVIID Paired Retrieval}
\label{app:aviid_transfer}

Table~\ref{tab:aviid_transfer_full} gives the full paired image-image retrieval behind the auxiliary AVIID sanity check summarized in the main paper. Two configurations are evaluated on the 804 AVIID test triplets: configuration~A retrieves between real infrared and visible images, and configuration~C retrieves between real infrared and DiffV2IR-synthesized infrared. In both configurations the IR-aware fine-tuned CLIP improves over its zero-shot counterpart in every direction and at every cutoff, with mean-recall gains of $+6.5$ (A) and $+23.4$ (C) points. Because the fine-tuning never uses AVIID data, this result provides limited auxiliary evidence of cross-dataset transfer under paired retrieval, rather than a claim of full real-sensor generalization.

\begin{table}[H]
  \centering
  \caption{Full AVIID paired image-image retrieval sanity check (\%) with OpenAI CLIP ViT-L/14, zero-shot vs.\ IR-aware fine-tuned, on 804 test triplets. Configuration~A: real-IR $\leftrightarrow$ visible; configuration~C: real-IR $\leftrightarrow$ synthetic-IR. mR averages the six recall values per configuration.}
  \label{tab:aviid_transfer_full}
  \scriptsize
  \setlength{\tabcolsep}{5pt}
  \renewcommand{\arraystretch}{1.0}
  \resizebox{0.8\textwidth}{!}{%
  \begin{tabular}{@{}cccccc@{}}
    \toprule
    Config. & Direction & R@1 & R@5 & R@10 & mR \\
    \midrule
    \multicolumn{6}{@{}c@{}}{\cellcolor{oursrowgreen}\textbf{Zero-shot (original pretrained weights)}} \\
    \multirow{2}{*}{A: real-IR $\leftrightarrow$ visible} & IR$\rightarrow$RGB & 2.74 & 12.44 & 19.78 & \multirow{2}{*}{12.17} \\
      & RGB$\rightarrow$IR & 3.73 & 13.68 & 20.65 & \\
    \multirow{2}{*}{C: real-IR $\leftrightarrow$ synthetic-IR} & IR$\rightarrow$Syn & 7.46 & 23.51 & 36.44 & \multirow{2}{*}{19.84} \\
      & Syn$\rightarrow$IR & 3.98 & 18.91 & 28.73 & \\
    \midrule
    \multicolumn{6}{@{}c@{}}{\cellcolor{oursrowgreen}\textbf{IR-aware fine-tuned}} \\
    \multirow{2}{*}{A: real-IR $\leftrightarrow$ visible} & IR$\rightarrow$RGB & 4.48 & 15.42 & 25.12 & \multirow{2}{*}{\textbf{18.70}} \\
      & RGB$\rightarrow$IR & 6.72 & 24.75 & 35.70 & \\
    \multirow{2}{*}{C: real-IR $\leftrightarrow$ synthetic-IR} & IR$\rightarrow$Syn & 21.02 & 50.87 & 62.56 & \multirow{2}{*}{\textbf{43.24}} \\
      & Syn$\rightarrow$IR & 18.41 & 46.89 & 59.70 & \\
    \bottomrule
  \end{tabular}}
\end{table}

This auxiliary result is included to make the synthetic-to-real boundary explicit. The improvement on AVIID indicates that the selected CLIP checkpoint is not merely memorizing the synthetic benchmark surface, but the experiment is still a small paired-retrieval check rather than a full real-sensor benchmark. We therefore use it as supporting evidence for representation transfer and keep the main claims tied to the controlled \dataset{} evaluation.

\subsection{Seed Stability}
\label{app:seed_stability}

Table~\ref{tab:seed_full} expands the seed-stability check of the main paper. OpenAI CLIP ViT-B/32 is fine-tuned under three random seeds with an otherwise identical recipe and evaluated on the 9{,}720-image filtered split. The mean recall is $11.88 \pm 0.02$ (sample standard deviation over seeds), suggesting that this backbone is stable under the tested recipe rather than driven by a single-run artifact.

\begin{table}[H]
  \centering
  \caption{Seed-stability check for OpenAI CLIP ViT-B/32 on the 9{,}720-image filtered split. I2T is IR$\rightarrow$IR-aware-text, T2I is the reverse; mR averages the six recall values.}
  \label{tab:seed_full}
  \scriptsize
  \setlength{\tabcolsep}{5pt}
  \renewcommand{\arraystretch}{1.0}
  \resizebox{\textwidth}{!}{%
  \begin{tabular}{@{}cccccccc@{}}
    \toprule
    Seed & I2T R@1 & I2T R@5 & I2T R@10 & T2I R@1 & T2I R@5 & T2I R@10 & mR \\
    \midrule
    1 & 4.73 & 12.13 & 17.49 & 5.28 & 13.02 & 18.76 & 11.90 \\
    2 & 4.69 & 12.25 & 17.52 & 5.35 & 12.95 & 18.58 & 11.89 \\
    3 & 4.56 & 12.25 & 17.48 & 5.35 & 12.97 & 18.53 & 11.86 \\
    \midrule
    \multicolumn{7}{@{}c@{}}{Mean $\pm$ std} & $11.88 \pm 0.02$ \\
    \bottomrule
  \end{tabular}}
\end{table}

The seed check is deliberately narrow. It controls for run-to-run variation on one representative CLIP backbone and confirms that the reported OpenAI CLIP ViT-B/32 result is not an obvious single-seed outlier. It does not replace a full multi-seed study over every backbone, which remains a larger-scale reproducibility extension.

\subsection{Per-Backbone VLM Lexical Diagnostics}
\label{app:per_backbone_vlm}

Table~\ref{tab:vlm_per_backbone} reports the five lexical diagnostics for all six \irvlm{} backbones on every evaluation task (500 held-out prompts per task; all prompts returned parseable text). The aggregate table in the main paper averages each column over these backbones. The per-backbone view exposes the trade-off discussed in the main text: on explanatory prompts (Infrared-cue explanation, Visual-evidence QA) the backbones that reach the highest lexical IR-cue rates, such as Qwen2.5-VL and H2RSVLM, also produce the longest answers and the highest overclaim rates, whereas InstructBLIP stays short and conservative with much lower IR-cue rates. No single backbone dominates all five diagnostics simultaneously, which is why the paper reports behavior-specific lexical metrics rather than one aggregate score.

\begin{table}[H]
  \centering
  \caption{Per-backbone \irvlm{} lexical diagnostics across the seven evaluation tasks (500 prompts each). IR-cue, RGB-color, overclaim, and class-hit are percentages; avg.\ words is response length. RGB-color and overclaim are lower-is-better; IR-cue and class-hit are higher-is-better. These metrics describe response wording and do not measure full semantic correctness.}
  \label{tab:vlm_per_backbone}
  \scriptsize
  \setlength{\tabcolsep}{5pt}
  \renewcommand{\arraystretch}{1.15}
  \resizebox{\textwidth}{!}{%
  \begin{tabular}{@{}ccccccc@{}}
    \toprule
    Task & Backbone & IR-cue $\uparrow$ & RGB-color $\downarrow$ & Overclaim $\downarrow$ & Class-hit $\uparrow$ & Avg.\ words \\
    \midrule
    \multirow{6}{*}{Captioning}
      & Qwen2.5-VL-7B & 100.0 & 0.0 & 0.4 & 0.8 & 17.9 \\
      & InstructBLIP-FLAN-T5-XL & 100.0 & 0.0 & 0.0 & 1.0 & 17.1 \\
      & LLaVA-1.5-7B & 100.0 & 0.0 & 0.0 & 1.0 & 17.8 \\
      & LLaVA-1.6-Vicuna-7B & 100.0 & 0.0 & 0.0 & 0.8 & 18.4 \\
      & GeoChat-7B & 100.0 & 0.0 & 0.0 & 0.8 & 17.9 \\
      & H2RSVLM-VHM-7B & 100.0 & 0.0 & 0.0 & 1.2 & 17.7 \\
    \midrule
    \multirow{6}{*}{Scene question}
      & Qwen2.5-VL-7B & 58.8 & 0.0 & 0.0 & 0.4 & 7.9 \\
      & InstructBLIP-FLAN-T5-XL & 12.4 & 0.0 & 0.0 & 0.2 & 2.4 \\
      & LLaVA-1.5-7B & 82.4 & 0.0 & 0.2 & 0.4 & 9.9 \\
      & LLaVA-1.6-Vicuna-7B & 88.2 & 0.0 & 0.2 & 0.2 & 11.0 \\
      & GeoChat-7B & 79.6 & 0.0 & 0.0 & 0.6 & 9.7 \\
      & H2RSVLM-VHM-7B & 66.4 & 0.0 & 0.0 & 0.6 & 7.4 \\
    \midrule
    \multirow{6}{*}{Object presence}
      & Qwen2.5-VL-7B & 0.4 & 0.0 & 0.0 & 0.8 & 3.3 \\
      & InstructBLIP-FLAN-T5-XL & 0.0 & 0.0 & 0.0 & 0.6 & 3.1 \\
      & LLaVA-1.5-7B & 0.0 & 0.0 & 0.0 & 0.6 & 3.5 \\
      & LLaVA-1.6-Vicuna-7B & 0.0 & 0.0 & 0.0 & 0.4 & 3.5 \\
      & GeoChat-7B & 0.4 & 0.0 & 0.0 & 0.4 & 3.6 \\
      & H2RSVLM-VHM-7B & 0.2 & 0.0 & 0.0 & 1.0 & 3.4 \\
    \midrule
    \multirow{6}{*}{Infrared-cue expl.}
      & Qwen2.5-VL-7B & 100.0 & 0.0 & 9.4 & 0.8 & 28.1 \\
      & InstructBLIP-FLAN-T5-XL & 100.0 & 0.0 & 0.0 & 0.4 & 12.8 \\
      & LLaVA-1.5-7B & 60.6 & 2.6 & 7.0 & 0.6 & 13.7 \\
      & LLaVA-1.6-Vicuna-7B & 79.2 & 2.0 & 12.0 & 0.2 & 15.4 \\
      & GeoChat-7B & 56.6 & 1.8 & 10.8 & 0.4 & 13.8 \\
      & H2RSVLM-VHM-7B & 100.0 & 2.8 & 24.8 & 1.2 & 45.9 \\
    \midrule
    \multirow{6}{*}{Hard scene}
      & Qwen2.5-VL-7B & 61.2 & 0.0 & 0.0 & 0.6 & 12.9 \\
      & InstructBLIP-FLAN-T5-XL & 0.0 & 0.0 & 0.0 & 0.2 & 1.2 \\
      & LLaVA-1.5-7B & 73.8 & 0.0 & 0.0 & 0.2 & 12.8 \\
      & LLaVA-1.6-Vicuna-7B & 64.4 & 0.0 & 0.0 & 0.6 & 14.5 \\
      & GeoChat-7B & 72.8 & 0.0 & 0.0 & 0.4 & 14.2 \\
      & H2RSVLM-VHM-7B & 58.4 & 0.0 & 0.0 & 0.6 & 13.4 \\
    \midrule
    \multirow{6}{*}{Hard-negative}
      & Qwen2.5-VL-7B & 26.6 & 1.4 & 0.0 & 0.4 & 15.6 \\
      & InstructBLIP-FLAN-T5-XL & 2.0 & 0.0 & 0.0 & 0.2 & 1.5 \\
      & LLaVA-1.5-7B & 65.8 & 3.2 & 0.0 & 1.0 & 18.1 \\
      & LLaVA-1.6-Vicuna-7B & 35.4 & 0.6 & 0.0 & 1.0 & 28.0 \\
      & GeoChat-7B & 46.6 & 1.4 & 0.0 & 1.6 & 30.2 \\
      & H2RSVLM-VHM-7B & 30.8 & 0.0 & 0.0 & 1.0 & 26.6 \\
    \midrule
    \multirow{6}{*}{Visual-evidence QA}
      & Qwen2.5-VL-7B & 100.0 & 0.0 & 33.4 & 1.4 & 35.1 \\
      & InstructBLIP-FLAN-T5-XL & 100.0 & 0.0 & 16.2 & 1.2 & 33.5 \\
      & LLaVA-1.5-7B & 26.0 & 0.0 & 8.8 & 0.4 & 10.0 \\
      & LLaVA-1.6-Vicuna-7B & 28.4 & 0.0 & 7.8 & 0.6 & 10.1 \\
      & GeoChat-7B & 27.4 & 0.2 & 7.6 & 0.4 & 9.6 \\
      & H2RSVLM-VHM-7B & 100.0 & 0.2 & 45.4 & 1.0 & 34.9 \\
    \bottomrule
  \end{tabular}}
\end{table}

The per-backbone table is intended as a diagnostic breakdown rather than a leaderboard. It shows that the selected VLM recipe makes captioning consistently use infrared-style language, while explanatory prompts remain more fragile because longer answers increase unsupported physical wording. The practical conclusion is therefore bounded: the fine-tuned VLMs are usable as first-version infrared instruction models, but their open-ended reasoning should still be interpreted through the lexical checks defined in Appendix~\ref{app:vlm_metrics}.

\clearpage
\section{Reproducibility Details}
\label{app:reproducibility}

This appendix reports the prompts, hyperparameters, and generation settings used to produce the dataset and all reported results. Values are transcribed from the release scripts.

\subsection{VLM Instruction and Evaluation Prompts}
\label{app:vlm_prompts}

The final API-v3 multitask recipe covers seven prompt types. Each training instruction is prefixed with the image token; the answers are rewritten by a text model and audited by a vision model before use. Table~\ref{tab:vlm_prompts} lists the evaluation prompts (500 held-out infrared images per task) verbatim.

\begin{table}[H]
  \centering
  \caption{Evaluation prompts for the seven \irvlm{} diagnostic tasks. Each row defines one fixed held-out prompt template used for VLM behavior analysis; the left column names the diagnostic task, and the right column gives the exact prompt issued to each model. The prompt set covers concise captioning, scene classification, object listing, evidence-grounded explanation, visual-evidence QA, hard scene discrimination, and hard-negative contrast under the same infrared remote-sensing image pool.}
  \label{tab:vlm_prompts}
  \scriptsize
  \setlength{\tabcolsep}{4pt}
  \setlength{\extrarowheight}{2pt}
  \renewcommand{\arraystretch}{1.22}
  \begin{tabular}{@{}>{\centering\arraybackslash}m{0.23\textwidth}>{\centering\arraybackslash}m{0.64\textwidth}@{}}
    \toprule
    Task & Prompt \\
    \midrule
    Infrared image captioning & Describe this infrared remote-sensing image in one concise sentence. \\
    \addlinespace[2pt]
    Scene classification & Classify the scene type in this infrared remote-sensing image. Answer with the scene name and one short visual reason. \\
    \addlinespace[2pt]
    Object presence & List the main visible objects or land-cover elements in this infrared remote-sensing image. \\
    \addlinespace[2pt]
    Infrared-cue explanation & Explain the infrared visual cues that support your interpretation of this image. \\
    \addlinespace[2pt]
    Visual-evidence QA & What visible infrared evidence supports the scene interpretation? \\
    \addlinespace[2pt]
    Hard scene discrimination & Choose the most likely scene class for this infrared image and explain why it is not a confusing alternative. \\
    \addlinespace[2pt]
    Hard-negative contrast & Describe one plausible wrong scene label for this image and explain why that label is less likely. \\
    \bottomrule
  \end{tabular}
\end{table}

These prompts define the evaluation interface rather than a hidden training advantage. Each task uses the same held-out image pool and a fixed prompt template, so changes in the diagnostic metrics reflect model behavior under comparable instructions. The prompt wording is included verbatim to make later reruns sensitive to the same instruction constraints.

\Needspace{0.42\textheight}
\subsection{IR-Aware Caption Generation}
\label{app:caption_prompt}

The caption-rewriting prompt is included as a compact card to make the supervision recipe auditable without expanding the main text.
\vspace{-0.6em}
\begin{figure}[H]
  \centering
  \includegraphics[width=0.86\linewidth]{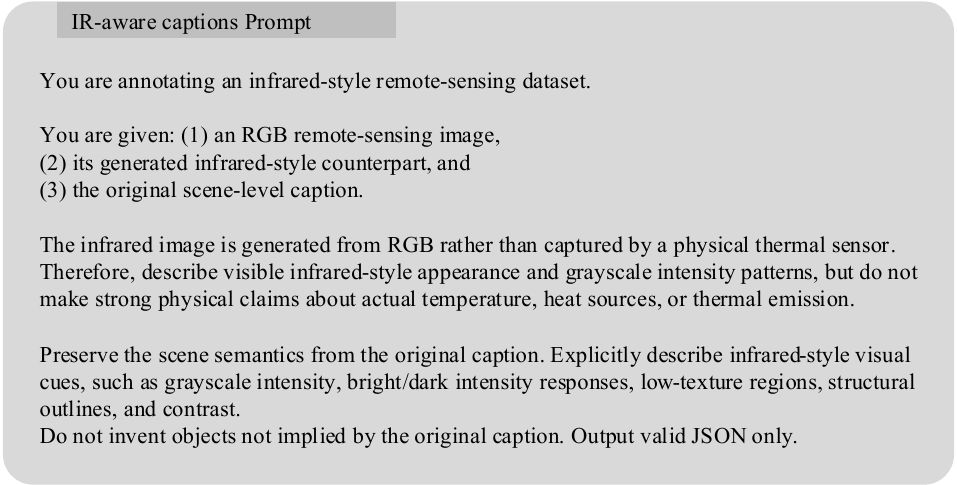}
  \caption{Prompt card for IR-aware caption generation.}
  \label{fig:ir_caption_prompt}
\end{figure}
\vspace{-0.4em}

\subsection{CLIP Fine-Tuning Hyperparameters}
\label{app:clip_hparams}

All \irclip{} backbones are fine-tuned with a frozen text encoder, AdamW, weight decay $0.1$, gradient clipping $1.0$, fp16 mixed precision, and a linear-warmup (20 steps) plus cosine-decay schedule. The input resolution follows each backbone's native CLIP processor ($224^2$). Per-backbone settings are listed in Table~\ref{tab:clip_hparams}; OpenAI ViT-L/14 additionally uses a stage-2 resume (6 epochs, lr $5\times10^{-7}$) from its stage-1 checkpoint.

\begin{table}[H]
  \centering
  \caption{\irclip{} fine-tuning hyperparameters. Batch sizes are the preferred values; the training queue automatically backs off under out-of-memory conditions.}
  \label{tab:clip_hparams}
  \footnotesize
  \setlength{\tabcolsep}{5pt}
  \renewcommand{\arraystretch}{1.1}
  \resizebox{0.8\textwidth}{!}{%
  \begin{tabular}{@{}ccccc@{}}
    \toprule
    Backbone & Epochs & Batch & LR & Text enc. \\
    \midrule
    OpenAI CLIP ViT-B/32 & 2 & 768 & $2\times10^{-6}$ & frozen \\
    OpenAI CLIP ViT-L/14 & 2 (+6) & 256 (128) & $1\times10^{-6}$ ($5\times10^{-7}$) & frozen \\
    OpenCLIP ViT-B/32 & 2 & 1024 & $2\times10^{-6}$ & frozen \\
    GeoRSCLIP ViT-B/32 & 2 & 1024 & $2\times10^{-6}$ & frozen \\
    RemoteCLIP ViT-B/32 & 2 & 1024 & $2\times10^{-6}$ & frozen \\
    \bottomrule
  \end{tabular}}
\end{table}

The CLIP configuration keeps the text side fixed and concentrates adaptation on the image side and projection parameters. This design makes the comparison across backbones easier to interpret because the language space is not re-learned for each model. The extra stage-2 resume for OpenAI CLIP ViT-L/14 is reported separately because it is the only CLIP run with an extended schedule.

\subsection{VLM LoRA Fine-Tuning Hyperparameters}
\label{app:vlm_hparams}

All six \irvlm{} backbones use frozen base language models and vision towers with LoRA adapters. Shared LoRA and optimizer settings are fixed across runs, while effective batch sizes and target modules differ by backbone, as summarized in Table~\ref{tab:vlm_hparams}.

\begin{table}[H]
  \centering
  \caption{\irvlm{} LoRA fine-tuning settings. Effective batch is per-device batch $\times$ gradient accumulation; all runs share $r{=}32$, $\alpha{=}64$, dropout $0.05$, lr $2\times10^{-4}$, and one epoch.}
  \label{tab:vlm_hparams}
  \footnotesize
  \setlength{\tabcolsep}{4pt}
  \renewcommand{\arraystretch}{1.1}
  \resizebox{\textwidth}{!}{%
  \begin{tabular}{@{}cccc@{}}
    \toprule
    Backbone & Eff.\ batch & Max len & LoRA target modules \\
    \midrule
    Qwen2.5-VL-7B & $1\times8$ & processor default & q,k,v,o,gate,up,down proj \\
    InstructBLIP-FLAN-T5-XL & $2\times8$ & 128 & q, v \\
    LLaVA-1.5-7B & $4\times4$ & 1024 & attention/MLP proj (LLaVA default) \\
    LLaVA-1.6-Vicuna-7B & $4\times4$ & 1024 & attention/MLP proj (LLaVA default) \\
    GeoChat-7B & $4\times4$ & 1024 & attention/MLP proj (LLaVA default) \\
    H2RSVLM-VHM-7B & $4\times4$ & 1024 & attention/MLP proj (LLaVA default) \\
    \bottomrule
  \end{tabular}}
\end{table}

This parameter-efficient setup isolates adapter and projector adaptation while keeping the six-backbone comparison controlled.

\subsection{DiffV2IR Generation Settings}
\label{app:diffv2ir}

Infrared images are synthesized with DiffV2IR using the \texttt{after\_phase\_2} checkpoint, a k-diffusion Euler-ancestral sampler, 20 steps, and $512$ long-side resolution aligned to multiples of 64. Generation uses three-way classifier-free guidance with text scale $7.5$, image and segmentation scales $1.5$, seed $1234$, and an Otsu-threshold segmentation map as the second condition. Outputs are saved as JPEG (quality 95) for all five sources and for the AVIID visible test images used in the realism study.

\subsection{Realism-Metric Computation}
\label{app:realism_compute}

FID uses torchvision Inception-v3 (\texttt{IMAGENET1K\_V1}) without the classifier head, yielding 2048-dimensional pooled features. Images are converted to grayscale, replicated to three channels, resized to $299^2$, and normalized with ImageNet statistics; FID is computed over exactly 804 aligned AVIID triplets (RGB / real-IR / synthetic-IR). Histogram distances use 256 normalized intensity bins over $[0,255]$ before computing the KL, JS, $\chi^2$, and intersection quantities of Appendix~\ref{app:ir_realism_metrics}.

\clearpage
\section{Extended Discussion}
\label{app:extended_discussion}

\medskip
\noindent\textbf{What the benchmark is designed to test.}
\dataset{} is intended to evaluate whether existing vision-language backbones can be redirected toward infrared remote-sensing evidence under a controlled data boundary. The benchmark is therefore not a general claim that synthetic imagery can replace real thermal sensors. Its narrower purpose is to isolate the language-alignment problem created by the infrared modality: images lose visible color cues, captions must stop referring to unsupported RGB appearance, and models must ground their responses in grayscale structure, intensity contrast, object-background separation, and scene layout. This scope explains why the paper evaluates both retrieval models and generative VLMs. CLIP tests whether infrared images and IR-aware captions can be aligned in a shared embedding space, while VLMs test whether instruction-following models can describe and reason about the same evidence without reverting to color-centric language.

\medskip
\noindent\textbf{Why IR-aware supervision matters.}
The central difficulty in infrared remote-sensing vision-language learning is not merely a visual domain shift. It is also a supervision shift. Captions written for visible-band images often describe color, illumination, and surface appearance that are absent or unreliable in infrared imagery. If such captions are reused without rewriting, the image encoder is asked to align grayscale intensity structure with language grounded in RGB evidence. The supervision ablation in the main paper supports this interpretation: replacing original captions with IR-aware captions raises development retrieval mean recall from 30.2\% to 43.0\% averaged across the five CLIP backbones. This gain indicates that the text side must describe the evidence actually visible to the model, such as intensity layout, object-background contrast, texture, and scene geometry, rather than treating infrared imagery as desaturated RGB.

\medskip
\noindent\textbf{How to interpret the retrieval scores.}
The absolute CLIP retrieval numbers remain modest even after fine-tuning, with the best model reaching 19.2\% mean recall on the filtered 9,720-image test split. This should be read as evidence that the task is difficult, not as failure of the adaptation recipe. Remote-sensing scenes often contain repeated layouts, visually similar land-cover patterns, and weak object-level discriminators, and the infrared modality removes many color shortcuts used by RGB-pretrained models. The important comparison is therefore against each backbone's own zero-shot behavior under the same candidate pool. All five CLIP backbones improve after IR-aware adaptation, with gains from $+3.5$ to $+12.8$ mean-recall points. The result shows that infrared-text alignment is learnable, while also leaving substantial headroom for future real-sensor data, larger-scale supervision, and stronger hard-negative training.

\medskip
\noindent\textbf{Why CLIP and VLM are evaluated separately.}
CLIP and VLM fine-tuning answer different questions. CLIP retrieval measures whether infrared images and IR-aware captions occupy a shared embedding space, so its main evidence is bidirectional recall on the filtered 9,720-image test split. VLM adaptation measures whether a generative model follows infrared instructions without reverting to visible-color language or unsupported thermal claims. This requires lexical behavior diagnostics rather than a single retrieval score. For captioning, the lexical IR-cue diagnostic reaches 100\% across all six backbones and RGB-color leakage falls to zero in the zero-shot comparison, but longer explanatory prompts still produce more overclaiming. This is why the paper reports task-separated diagnostics: frequent infrared wording, low color leakage, and semantic correctness are related but not identical properties.

\medskip
\noindent\textbf{Why synthetic infrared remains useful.}
The image pool is synthesized because large-scale paired infrared remote-sensing image-text data are not available at the scale required for CLIP and VLM adaptation. This choice introduces a clear boundary: the generated images should be viewed as an infrared-style training and evaluation surface, not as radiometrically calibrated sensor measurements. However, the synthetic modality is not equivalent to simple grayscale conversion. On AVIID, DiffV2IR-based synthesis is closer to real infrared than RGB-to-gray under FID and grayscale-histogram distances, and the selected IR-CLIP checkpoint also improves auxiliary paired retrieval with real-infrared queries. These checks do not prove full sensor realism, but they indicate a nontrivial infrared-style alignment signal beyond simple grayscale conversion under the reported proxy metrics.

\medskip
\noindent\textbf{Why RGB--IR dual-modal learning is not the main target here.}
The construction pipeline starts from visible remote-sensing images, but the model-facing task deliberately exposes the infrared image and IR-aware text. This choice separates the paper from RGB--IR matching or fusion benchmarks. A dual-modal objective may improve paired cross-modal retrieval, but it can also let models depend on visible-domain semantics that are not present at inference time when only infrared imagery is provided. The ablation in the main paper reflects this trade-off: joint RGB--IR diagnostic training obtains a similar IR-aware development score, yet the selected recipe prioritizes infrared-text alignment because it matches the intended model input. This does not imply that RGB--IR learning is unimportant; rather, it is a different problem with a different deployment assumption.

\medskip
\noindent\textbf{What the VLM diagnostics do and do not prove.}
The VLM diagnostics are intentionally conservative lexical checks. They verify whether generated answers mention infrared evidence, avoid visible-color leakage, avoid unsupported physical claims, and sometimes repeat a source class token. These measurements are useful because the dataset does not contain human-verified dense semantic labels for every image. They should not be interpreted as full object recognition accuracy or expert-level scene understanding. For example, a high IR-cue rate means the model uses infrared-style evidence language, while a low RGB-color rate means it avoids one common failure mode. The next step is to add human-verified answers or external semantic labels so that lexical behavior can be connected to correctness.

\medskip
\noindent\textbf{Limitations and future directions.}
The main limitations follow directly from the data construction. First, the infrared images are generated from visible sources, so the benchmark cannot replace real sensor-captured infrared evaluation. Second, IR-aware captions and instruction answers are produced by strong foundation models and filtered automatically, which makes the supervision scalable but not equivalent to expert annotation. Third, most CLIP and VLM numbers are single-run measurements, except for the OpenAI CLIP ViT-B/32 seed check. Future work should add real infrared sensor data, radiometric metadata where available, human-verified instruction answers, larger multi-seed studies, stronger hard-negative retrieval, and semantic evaluation beyond lexical proxy metrics.

\clearpage
\section{Qualitative Case Gallery and Failure Cases}
\label{app:qualitative_failures}

The qualitative analysis follows an extended-appendix case-table style: each example records the task setting, infrared image, prompt or query, model output, and diagnostic label. This layout makes the examples auditable rather than purely illustrative.

\medskip
\noindent\textbf{CLIP retrieval cases.}
For CLIP, the cases cover successful and partial retrieval, semantic near misses, and hard visual neighbors in repetitive infrared layouts. Successful examples align distinctive global geometry with IR-aware captions describing scene structure, including circular fields, runways, dense building blocks, road grids, water boundaries, and industrial tanks. Near misses arise when classes share similar infrared geometry, such as industrial areas versus dense residential blocks, river boundaries versus dark road corridors, or rail stations versus other linear transportation scenes. Because infrared imagery removes color cues and many discriminative objects occupy small regions, these examples reveal both learned alignment and residual candidate-pool ambiguity.

\vspace{-2.0em}
\begin{table}[H]
  \centering
  \caption{Qualitative CLIP retrieval display cases. Each row restores the case-table view used for manual inspection: the held-out query image, the target caption, the retrieved evidence, and the diagnostic reading are shown together. We include eight representative cases to cover correct retrieval, partial retrieval, semantic near misses, and hard visual neighbors.}
  \label{tab:qual_clip_cases}
  \tiny
  \setlength{\tabcolsep}{3pt}
  \renewcommand{\arraystretch}{1.06}
  \resizebox{\textwidth}{!}{%
  \begin{tabular}{@{}C{0.14\textwidth}C{0.13\textwidth}C{0.27\textwidth}G{0.29\textwidth}C{0.17\textwidth}@{}}
    \toprule
    Case & Query image & Target caption & Retrieved evidence & Diagnostic \\
    \midrule
    Correct top-1 &
    \includegraphics[width=0.82\linewidth]{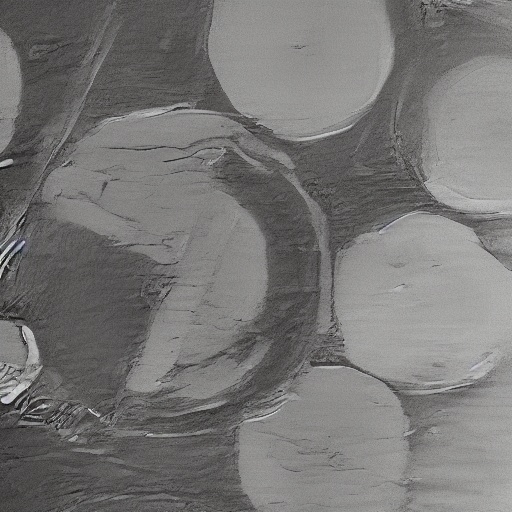} &
    Circular farmlands appear as bright rounded regions against a darker, low-texture background. &
    Rank 1 retrieves the paired circular-farmland caption; lower ranks contain related circular storage or industrial layouts. &
    Correct match at rank 1; global circular geometry is aligned. \\
    Correct in top-5 &
    \includegraphics[width=0.82\linewidth]{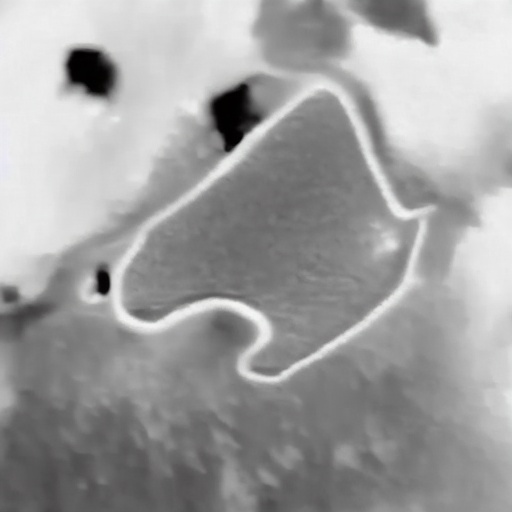} &
    A bright island boundary contrasts with darker surrounding water and textured vegetation. &
    Rank 1 is a wetland with dark water boundaries; rank 2 is the paired island caption. &
    Correct match appears after a visually close water-boundary neighbor. \\
    Semantic near miss &
    \includegraphics[width=0.82\linewidth]{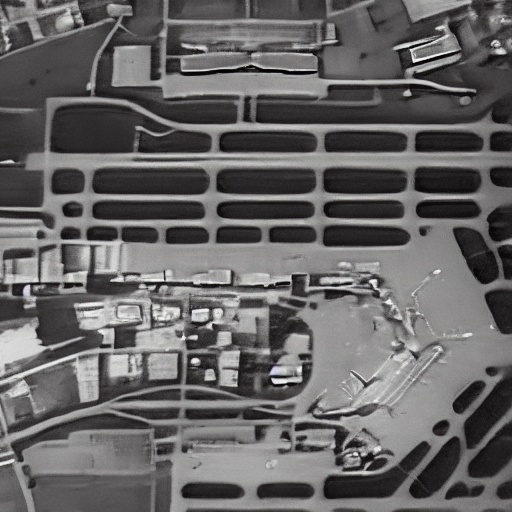} &
    An airport-like grid of runways and bright building regions appears against darker surroundings. &
    Top retrieved captions describe airport layouts with runways, taxiways, and strong grayscale contrast. &
    Airport semantics are retrieved, but the exact pair falls outside top-10. \\
    Hard visual neighbor &
    \includegraphics[width=0.82\linewidth]{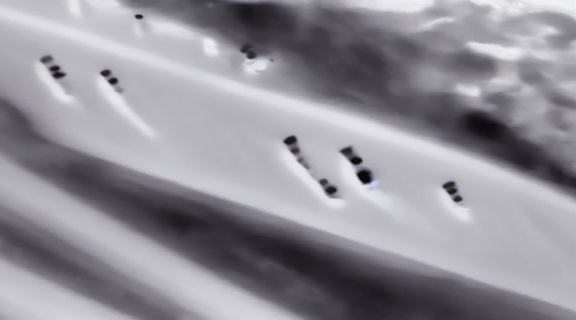} &
    Parking-like bright compact objects are arranged over a darker paved surface with regular boundaries. &
    Top evidence mixes dense built-up pavement, parking-lot captions, and industrial-yard layouts. &
    Repeated paved layouts create close competitors. \\
    Baseball-field match &
    \includegraphics[width=0.82\linewidth]{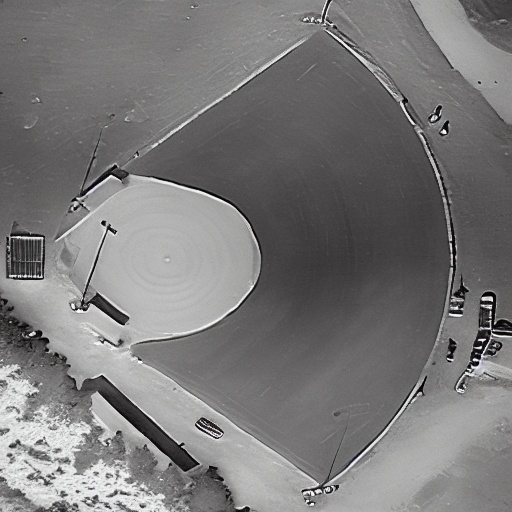} &
    A baseball field is visible as a broad rounded infield and darker surrounding outfield. &
    Retrieved evidence describes a baseball diamond and adjacent low-texture field boundaries. &
    Sports-field geometry gives a stable retrieval cue. \\
    Oil-tank match &
    \includegraphics[width=0.82\linewidth]{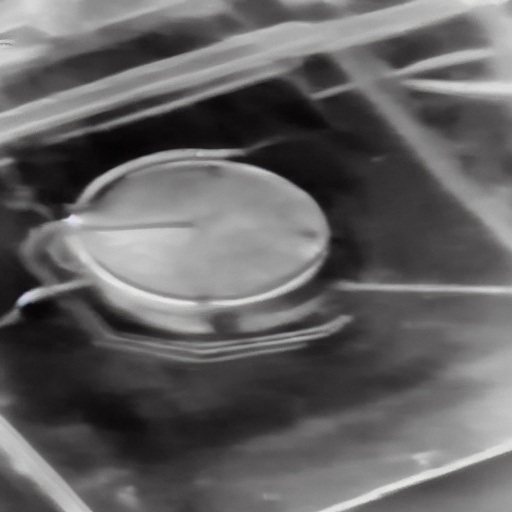} &
    A circular storage tank appears as a bright rounded structure beside linear road patterns. &
    Top captions emphasize circular tanks, industrial surfaces, and adjacent access roads. &
    Circular industrial objects are visually distinctive. \\
    Rail-station neighbor &
    \includegraphics[width=0.82\linewidth]{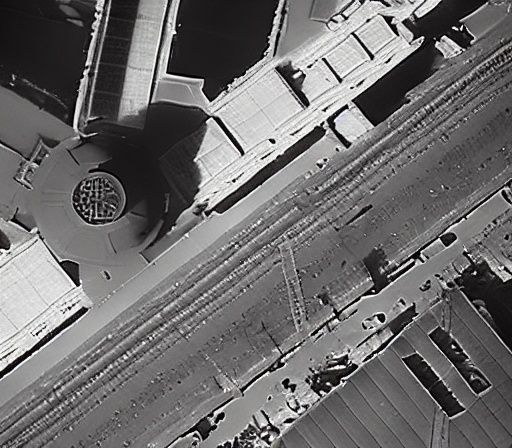} &
    Parallel rail tracks and station-like roof structures form a dense transportation layout. &
    Retrieved evidence mixes train-station, bridge, and dense transport-infrastructure captions. &
    Linear infrastructure creates plausible near neighbors. \\
    Reservoir boundary &
    \includegraphics[width=0.82\linewidth]{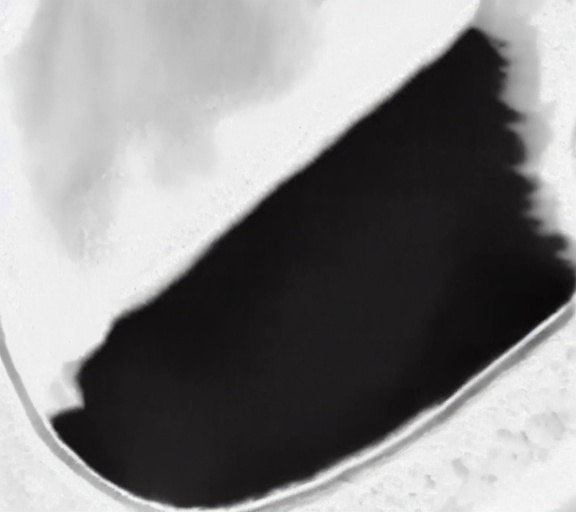} &
    A dark water body is bounded by a bright shoreline and smooth surrounding terrain. &
    Top evidence favors reservoir or lake captions with strong water-land contrast. &
    Water-boundary contrast supports alignment. \\
    \bottomrule
  \end{tabular}}
\end{table}

\medskip
\noindent\textbf{VLM generation cases.}
For VLMs, examples are grouped by prompt type because captioning, classification, object listing, evidence QA, cue explanation, hard-scene discrimination, hard-negative contrast, and color-leakage checking stress different behaviors. A useful response should identify the remote-sensing scene, ground the answer in visible infrared structures, and avoid importing unsupported visible-color descriptions. The successful cases typically mention grayscale contrast, bright-dark boundaries, compact object arrangements, linear transportation structures, or water-land separation as evidence. The more difficult cases test whether the model can distinguish visually similar layouts, such as airports versus roads, rail stations versus other dense infrastructure, bridges versus linear banks, or reservoirs versus dark background regions. These examples therefore complement the quantitative diagnostics by showing whether the generated answers are concise, visually grounded, and consistent with the intended modality.

\vspace{-2.0em}
\begin{table}[H]
  \centering
  \caption{Qualitative VLM generation display cases. Each row shows the held-out image, the evaluation prompt, the generated answer, and the manual diagnostic label used when preparing the case gallery. We include eight prompt types to show concise captioning, classification, object listing, evidence-grounded QA, cue explanation, hard-scene discrimination, hard-negative contrast, and color-leakage checking.}
  \label{tab:qual_vlm_cases}
  \tiny
  \setlength{\tabcolsep}{3pt}
  \renewcommand{\arraystretch}{1.06}
  \resizebox{\textwidth}{!}{%
  \begin{tabular}{@{}C{0.15\textwidth}C{0.13\textwidth}C{0.24\textwidth}G{0.31\textwidth}C{0.17\textwidth}@{}}
    \toprule
    Task & Image & Prompt & Generated answer & Diagnostic \\
    \midrule
    Captioning &
    \includegraphics[width=0.82\linewidth]{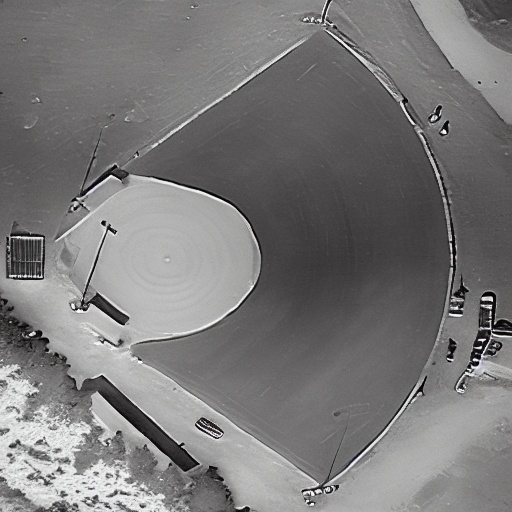} &
    Describe this infrared image in one sentence. &
    A baseball field shows a bright infield, darker outfield, and low-texture surroundings. &
    Concise scene grounding. \\
    Scene classification &
    \includegraphics[width=0.82\linewidth]{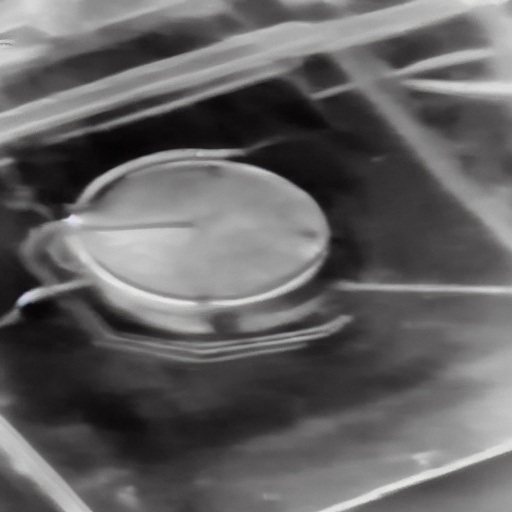} &
    Classify the scene and give one visual reason. &
    Industrial area; circular tank structure visible in grayscale. &
    Class plus visual reason. \\
    Infrared-cue explanation &
    \includegraphics[width=0.82\linewidth]{figures/cases/assets/vlm_cue_parking.jpg} &
    Explain the infrared visual cues. &
    High-contrast outlines and grayscale intensity variation reveal parked cars against darker pavement. &
    Uses intensity cues. \\
    Hard-negative contrast &
    \includegraphics[width=0.82\linewidth]{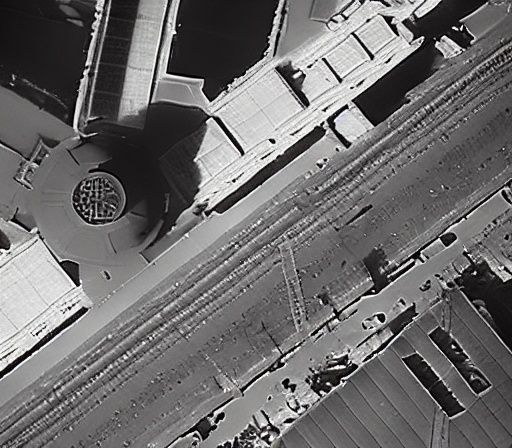} &
    Give a wrong label and explain why. &
    Airport; dense buildings and railway tracks are visible, while runways or terminals are absent. &
    Rejects confusing label. \\
    Object presence &
    \includegraphics[width=0.82\linewidth]{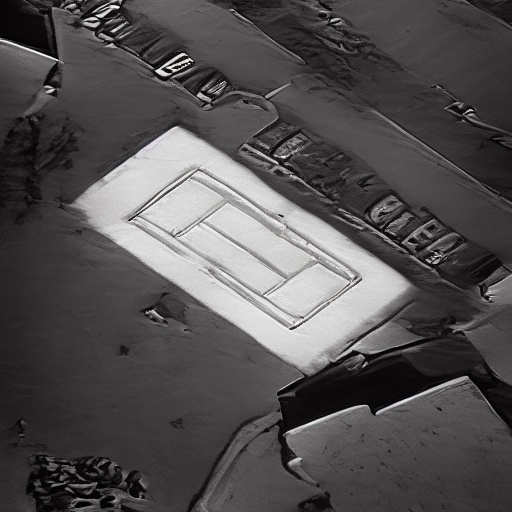} &
    List the main visible objects or land-cover elements. &
    Tennis courts, paved surfaces, nearby buildings, and dark boundary regions are visible. &
    Object list grounded in structure. \\
    Visual-evidence QA &
    \includegraphics[width=0.82\linewidth]{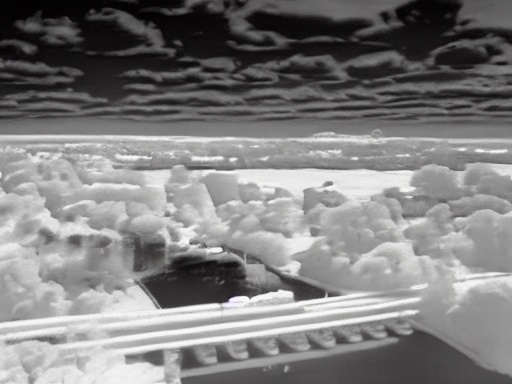} &
    What visible infrared evidence supports the scene interpretation? &
    The bright bridge deck, dark water channel, and repeated linear supports indicate a bridge scene. &
    Uses local visual evidence. \\
    Hard scene discrimination &
    \includegraphics[width=0.82\linewidth]{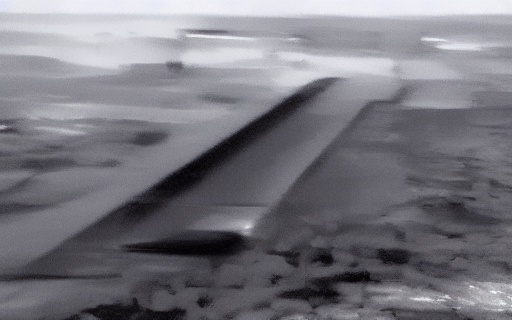} &
    Choose the most likely scene class and explain why it is not a confusing alternative. &
    Airport; the long runway-like strip and open surrounding area are more consistent than a road. &
    Distinguishes similar linear scenes. \\
    Color-leakage check &
    \includegraphics[width=0.82\linewidth]{figures/cases/assets/extra_reservoir.jpg} &
    Describe the image without using visible-color terms. &
    A dark reservoir occupies the center, bordered by bright shoreline and smoother surrounding terrain. &
    Avoids RGB-color wording. \\
    \bottomrule
  \end{tabular}}
\end{table}
\vspace{-2.0em}

\end{document}